\documentclass[10pt,journal,compsoc]{IEEEtran}
\ifCLASSOPTIONcompsoc
  \usepackage[nocompress]{cite}
\else
  \usepackage{cite}
\fi
\ifCLASSINFOpdf
\else
\fi

\usepackage{xspace}

\makeatletter
\DeclareRobustCommand\onedot{\futurelet\@let@token\@onedot}
\def\@onedot{\ifx\@let@token.\else.\null\fi\xspace}

\def\eg{\emph{e.g}\onedot} 
\def\ie{\emph{i.e}\onedot}

\makeatother

\usepackage{times}
\usepackage{epsfig}
\usepackage{graphicx}
\usepackage{amsmath}
\usepackage{amssymb}

\usepackage{array}
\usepackage{epstopdf, booktabs}
\usepackage{enumitem}
\usepackage{subfigure}
\usepackage[english]{babel}
\usepackage{multirow}
\usepackage{lipsum}
\usepackage[linesnumbered,ruled]{algorithm2e}

\DeclareMathAlphabet{\mathpzc}{OT1}{pzc}{m}{it} % \mathcal small letters 

\newcolumntype{P}[1]{>{\centering\arraybackslash}p{#1}}
\newcolumntype{M}[1]{>{\centering\arraybackslash}m{#1}}

\newcommand{\tabincell}[2]{\begin{tabular}{@{}#1@{}}#2\end{tabular}}
\usepackage[pagebackref=true,breaklinks=true,letterpaper=true,colorlinks,bookmarks=false]{hyperref}

\begin{document}
%\title{BlockQNN: Practical Block-wise Neural Network Architecture Generation}
\title{BlockQNN: Efficient Block-wise Neural Network Architecture Generation}


\author{Zhao~Zhong,
        Zichen Yang, Boyang Deng, Junjie Yan, Wei Wu, Jing Shao,
        and~Cheng-Lin~Liu,~\IEEEmembership{Fellow,~IEEE}% <-this % stops a space
\IEEEcompsocitemizethanks{\IEEEcompsocthanksitem Z. Zhong is with the NLPR, Institute of Automation of Chinese Academy of Sciences, University of Chinese Academy of Sciences, Beijing 100190, P.R. China. E-mail: zhao.zhong@nlpr.ia.ac.cn.

\IEEEcompsocthanksitem Z. Yang, B. Deng, J. Yan, W. Wu and J. Shao are with Sensetime Research Institute.E-mail: \{yangzichen, dengboyang, yanjunjie, wuwei, shaojing\}@sensetime.com

\IEEEcompsocthanksitem C.-L. Liu is with the NLPR, Institute of Automation of Chinese Academy of Sciences, Beijing, China, and the CAS Center for Excellence in Brain Science and Intelligence Technology, University of Chinese Academy of Sciences, Beijing 100190, P.R. China. E-mail: liucl@nlpr.ia.ac.cn. 
}% <-this % stops an unwanted space

}
\IEEEtitleabstractindextext{%
\begin{abstract}
Convolutional neural networks have gained a remarkable success in computer vision.
However, most usable network architectures are hand-crafted and usually require expertise and elaborate design.
In this paper, we provide a block-wise network generation pipeline called BlockQNN which automatically builds high-performance networks using the Q-Learning paradigm with epsilon-greedy exploration strategy.
The optimal network block is constructed by the learning agent which is trained to choose component layers sequentially. We stack the block to construct the whole auto-generated network.
To accelerate the generation process, we also propose a distributed asynchronous framework and an early stop strategy.
The block-wise generation brings unique advantages:
(1) it yields state-of-the-art results in comparison to the hand-crafted networks on image classification, particularly, the best network generated by BlockQNN achieves $2.35\%$ top-$1$ error rate on CIFAR-$10$.
%, beating all existing auto-generated networks.
% 
(2) it offers tremendous reduction of the search space in designing networks, spending only $3$ days with $32$ GPUs.
A faster version can yield a comparable result with only 1 GPU in 20 hours.
%, which is affordable for common deep learning researcher.
% 
(3) it has strong generalizability in that the network built on CIFAR also performs well on the larger-scale dataset. The best network achieves very competitive accuracy of $82.0$\% top-$1$ and $96.0$\% top-5 on ImageNet.
\end{abstract}

}

\maketitle

\IEEEdisplaynontitleabstractindextext

% \fi
\IEEEpeerreviewmaketitle

\IEEEraisesectionheading{\section{Introduction}\label{sec:introduction}}

% ======= Background
\IEEEPARstart{D}{uring} the last decades, Convolutional Neural Networks (CNNs) have shown remarkable potentials almost in every field in the computer vision society~\cite{lecun2015deep}. 
It achieved successes first in image classification~\cite{krizhevsky2012imagenet}, and then in object detection~\cite{girshick2015fast,ren2015faster}, semantic segmentation~\cite{long2015fully,chen2018deeplab} and tracking~\cite{nam2016learning,bertinetto2016fully}.
For example, the network evolution from AlexNet~\cite{krizhevsky2012imagenet}, VGG~\cite{simonyan2014very}, Inception~\cite{szegedy2015going} to ResNet~\cite{he2015deep} has improved the top-5 performance on ImageNet challenge steadily from $83.6\%$ to $96.43\%$. 
However, as the performance gain usually requires an increasing network capacity, a high-performance network architecture generally possesses a tremendous number of possible configurations about the number of layers, hyperparameters in each layer and type of each layer.
It is hence infeasible to find the optimal network structure by manually exhaustive search, and the design of successful hand-crafted networks heavily rely on expert knowledge and experience.
Therefore, constructing network in a smart and automatic manner remains an open problem. 

%=========== fig: fig1 ==================
\begin{figure*}[tbp]
	\centering
	\includegraphics[width=1\textwidth]{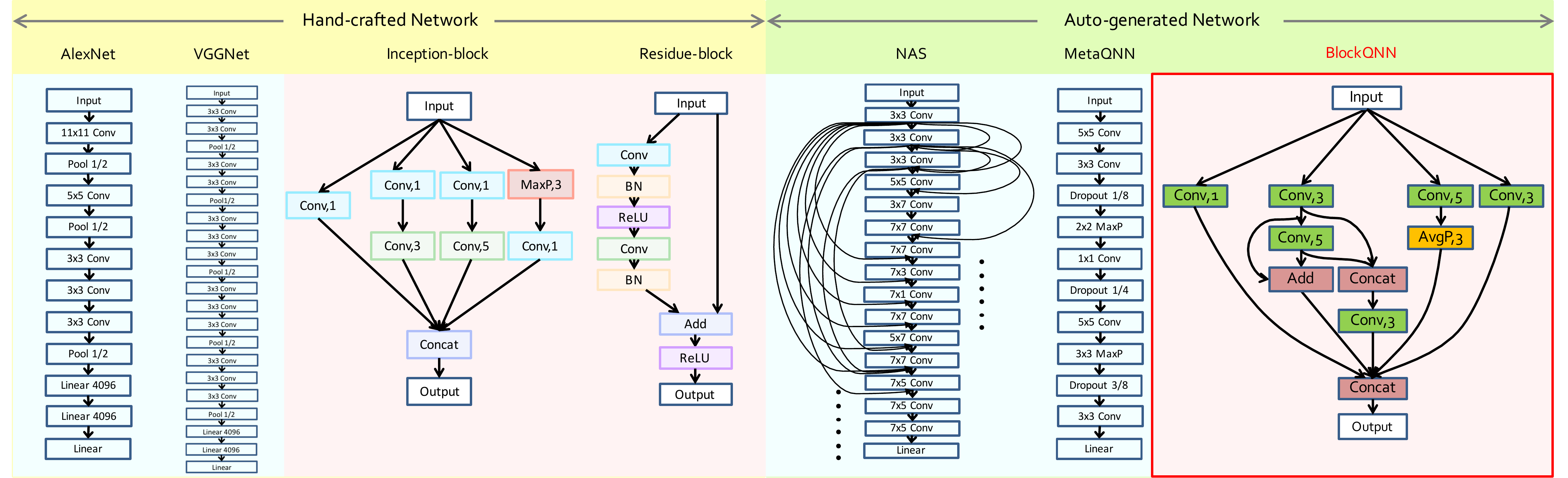}
	\caption{The proposed \textbf{BlockQNN} (right in red box) compared with the hand-crafted networks marked in yellow and the existing auto-generated networks in green. Automatically generating the \texttt{plain} networks~\cite{baker2016designing,zoph2016neural} marked in blue need large computational costs in searching optimal layer types and hyperparameters for each single layer, while the \texttt{block}-wise network heavily reduces the cost to search structures only for one block. The entire network is then constructed by stacking the generated blocks. Similar \texttt{block} concept has been demonstrated its superiority in hand-crafted networks, such as inception-block and residue-block marked in red.}
	%\vspace{-0.1cm}
	\label{fig:motivation_compare}
\end{figure*}
%=========== fig: fig1 ==================

% ========= Challenge of automatic model generation
Although some recent works have attempted computer-aided or automated network design~\cite{baker2016designing,zoph2016neural}, there are several challenges unsolved: 
(1) The large number of convolutional layers and the numerous options in type and hyperparameters of each make huge search space and heavy computational costs for network generation.
(2) The network designed on a specific dataset or task yields inferior performance when transfered to other datasets or tasks.
%, and thus is hard to transfer to other tasks or generalize to another dataset with different input data sizes.
% 
In this paper, we aim to solve the aforementioned challenges by proposing a novel fast Q-learning framework, called \textit{BlockQNN}, for automatically designing the network architecture, as shown in Fig.~\ref{fig:motivation_compare}.

% ====== Method point1
Particularly, to make the network generation efficient and generalizable, we generate the network in \texttt{block}-wise,~\ie, stacking personalized blocks rather than tedious per-layer network piling.
This ia inspired by some modern CNN architectures such as Inception~\cite{szegedy2015going,ioffe2015batch,szegedy2015rethinking} and ResNet Series~\cite{he2015deep,he2016identity} which are assembled as the stack of basic \texttt{block} structures.
For example, the inception and residual blocks shown in Fig.~\ref{fig:motivation_compare} are repeatedly concatenated to construct the entire network.
With such kind of \texttt{block}-wise network architecture, the generated network owns a powerful generalization to other task domains or different datasets.

% In this paper, we make efforts to design a CNN architecture in an automatic manner with a block-wise concept, efficiently and practically.

% ====== Method point2
% To overcome the greatest challenge of the huge demand of computing resource for this task, we focus on designing network blocks for building network, instead of generating whole network directly, to reduce the search space of network design.
% 
In comparison to previous methods like NAS~\cite{zoph2016neural} and MetaQNN~\cite{baker2016designing}, as depicted in Fig.~\ref{fig:motivation_compare}, we present a more readily and elegant model generation method that specifically designed for block-wise generation.
Motivated by the unsupervised reinforcement learning paradigm, we employ the well-known Q-learning~\cite{watkins1989learning} with experience replay~\cite{lin1993reinforcement} and epsilon-greedy strategy~\cite{mnih2015human} to effectively and efficiently search for the optimal block structure.
The network block is constructed by the learning agent which is trained sequentiality to choose component layers. Afterwards we stack the block to construct the whole auto-generated network.
%
%and stacking configuration.
% 
%Moreover, to enable efficient search with fast convergence, we also propose an early stop strategy with a new reward function aiming at balancing the network performance and the network complexity.
Moreover, we propose an early stop strategy to enable efficient search with fast convergence. A novel reward function is designed to ensure the accuracy of the early stopped network to be positively correlated with the converged network. Good blocks can be selected in reduced training time using this property.
With this acceleration strategy, we can construct a Q-learning agent to learn the optimal block-wise network structure for a given task with limited resources (\eg~few GPUs or short time period).
The generated architectures are thus succinct and have powerful generalization ability compared to the networks generated by the other automatic network generation methods.

%--new--The followed works
A preliminary version of this manuscript was published previously in conference~\cite{zhong2018blockqnn}. Since then, the block conception for auto-generated network have been adopted and generalized to other methods, such as~\cite{liu2017hierarchical,real2018regularized,pham2018faster,liu2017progressive,cai2018path}. Searching transferable blocks (referred to as cells in~\cite{zoph2018learning}) and assembling them into a network becomes an universal formulation in automating neural network design area.

%--we add some more
After that, we introduce more advanced depthwise convolution operation~\cite{chollet2016xception} to update the state-of-the-art performance on image classification, and in this paper we analyze the different connection styles between blocks instead of stacking block structures sequentially. Also, we propose the accelerated block-wise network generation with network performance prediction, called faster BlockQNN, which only costs 20 hours with 1 GPU on CIFAR.

% ========= Contribution
The proposed \texttt{block}-wise network generation method brings a few advantages as follows:
\begin{itemize}
	%\vspace{-0.2cm}
	\item \textit{Effectiveness}. The automatically generated networks present state-of-the-art performance compare to those of hand-crafted networks with human expertise.
	The proposed method is also superior to the existing automatic works and achieves a leading performance on CIFAR-$10$ with $2.35\%$ error rate.
	%\vspace{-0.25cm}
	\item \textit{Efficiency}. We are the first to consider \texttt{block}-wise setup in automatic network generation.
	The block-wise setup and the proposed early stop strategy result in a fast search process.
	The network generation for CIFAR task reaches convergence with only $32$ GPUs in $3$ days, which is much more efficient than that by NAS~\cite{zoph2016neural} with $800$ GPUs in $28$ days.
	Moreover, the faster version can get a comparable result with only 1 GPU in 20 hours, which is affordable for common deep learning researchers.
	%\vspace{-0.25cm}
	\item \textit{Transferability}. The proposed method offers surprisingly superior transferable ability that the network generated for CIFAR can be transferred to ImageNet with outstanding performance with little modification. The best network achieves very competitive accuracy of $82.0$\% top-$1$ and $96.0$\% top-5 on ImageNet.
\end{itemize}
%-------------------------------------------------------------------------


%%%%%%%%%%%%%%%%%%%%%%%%%%%%%%%%%%% RELATED WORK
\section{Related Work}
Early works, from $1980$s, have made efforts on automating neural network design which often searched good architecture by the genetic algorithm or other evolutionary algorithms~\cite{schaffer1992combinations,stanley2002evolving,stanley2009hypercube,suganuma2017genetic,saxena2016convolutional,domhan2015speeding,xie2017genetic}. Nevertheless, these works, to our best knowledge, cannot perform competitively compared with hand-crafted networks. 
% 
% Recent works, \ie~Neural Architecture Search (NAS)~\cite{zoph2016neural} and MetaQNN~\cite{baker2016designing}, based on reinforcement learning, provide powerful capability to beat the state-of-the-art hand-crafted networks.
% 
Recent works, \ie~Neural Architecture Search (NAS)~\cite{zoph2016neural} and MetaQNN~\cite{baker2016designing}, adopted reinforcement learning to automatically search a good network architecture. 
Although they can yield good performance on small datasets such as CIFAR-$10$, CIFAR-$100$, the direct use of MetaQNN or NAS for architecture design on big datasets like ImageNet~\cite{deng2009imagenet} is computationally expensive via searching in a huge space.
Besides, the network generated by this kind of methods is task-specific or dataset-specific, that is, it cannot been well transferred to other tasks nor datasets with different input data sizes. For example, the network designed for CIFAR-$10$ cannot been generalized to ImageNet.

Instead, our approach is aimed for designing network block architecture by an efficient search method with a distributed asynchronous Q-learning framework as well as an early-stop strategy.
The block design conception follows the modern convolutional neural networks such as Inception~\cite{szegedy2015going,ioffe2015batch,szegedy2015rethinking} and Resnet~\cite{he2015deep,he2016identity}. The inception-based networks construct the \texttt{inception blocks} via a hand-crafted multi-level feature extractor strategy by computing $1\times 1$, $3\times 3$, and $5\times 5$ convolutions, while the Resnet uses \texttt{residue blocks} with shortcut connection to make it easier to represent the identity mapping which allows a very deep network. 
The \texttt{blocks} automatically generated by our approach have similar structures such as some blocks contain short cut connections and inception-like multi-branch combination. We will discuss the details in Section~\ref{subsec:block_analysis}.
Concurrent with our work, the NASNet~\cite{zoph2018learning} is developed for learning block structures (referred to as cells in~\cite{zoph2018learning}) to construct the whole network with RNN-based controller.

There is a growing interest in improving the efficiency of automatic network generation for Common researchers who have limited computing resources. Baker et al.~\cite{baker2017practical} use standard frequentist regression models to predict the final performance,  Brock et al.~\cite{brock2017smash} propose SMASH which designs an architecture and then uses a hyper-network to generate its weights. These methods, however, can not compete with state-of-the-art networks, and the networks generated by them are still task-specific or dataset-specific.

Other related works include hyper-parameter optimization~\cite{bergstra2011algorithms}, meta-learning~\cite{vilalta2002perspective} and learning to learn methods~\cite{hochreiter2001learning,andrychowicz2016learning}. The goal of these works is to use meta-data to improve the performance of the existing algorithms, such as finding the optimal learning rate of optimization methods or the optimal number of hidden layers to construct the network. In this paper, we focus on learning the entire topological structure of network blocks to improve the performance.

%$t\in\mathbf{T}, (\mathbf{T}=\{1,2,3,...\})$
%$k,k^{'}\in \mathbf{K}, \mathbf{K}=\{1,2,...,t-1\}$.
%table 1
\begin{table}[t!]
	%% increase table row spacing, adjust to taste
	%\setlength{\tabcolsep}{3pt}
	\renewcommand{\arraystretch}{1.3}
	%\setlength{\belowcaptionskip}{3pt}
	%\vspace{-0.1cm}
	%\centering
	\begin{center}
		\caption{Network Structure Code Space. The space contains seven types of commonly used layers. Layer index stands for the position of the current layer in a block, the range of the parameters is set to be $\mathbf{T}=\{1,2,3,...\mbox{max layer index}\}$. Three kinds of kernel sizes are considered for convolution layer and two sizes for pooling layer. Pred$1$ and Pred$2$ refer to the predecessor parameters which is used to represent the index of preceding layers, the allowed range is $\mathbf{K}=\{1,2,...,\mbox{current layer index}-1\}$} \label{table:1}
		\footnotesize
		\begin{tabular}{c|c|c|c|c|c}
			\hline
			Name&Index&Type&Kernel Size&Pred1&Pred2 \\
			\hline
			Convolution&\(\mathbf{T}\)&1& 1, 3, 5 & \(\mathbf{K}\) & 0 \\
			\hline
			Max Pooling&\(\mathbf{T}\)&2&1, 3 & \(\mathbf{K}\) & 0 \\
			\hline
			Average Pooling&\(\mathbf{T}\)&3& 1, 3 & \(\mathbf{K}\) & 0\\
			\hline
			Identity&\(\mathbf{T}\)&4& 0 & \(\mathbf{K}\) & 0\\
			\hline
			Elemental Add&\(\mathbf{T}\)&5& 0 & \(\mathbf{K}\) & \(\mathbf{K}\)\\
			\hline
			Concat&\(\mathbf{T}\)&6& 0 & \(\mathbf{K}\) & \(\mathbf{K}\)\\
			\hline
			Terminal&\(\mathbf{T}\)&7& 0 & 0 & 0\\
			\hline
		\end{tabular}
	\end{center}

	%\vspace{-0.1cm}
\end{table}

%%%%%%%%%%%%%%%%%%%%%%%%%%%%%%%%%%% METHODOLOGY
\section{Methodology}
\label{sec:method}
In this section, we first present the basic designs and properties of the proposed BlockQNN framework. Extension of the framework to block connection style will be described in Section~\ref{subsec:block_conn}. The faster version with network performance prediction called faster BlockQNN will be introduced in Section~\ref{subsec:predict_network_prf} .

\subsection{ Convolutional Neural Network Blocks}
\label{subsec:blocks}

The modern CNN architectures, \eg~Inception and Resnet, are designed by stacking several \texttt{blocks} each of which shares similar structure but with different weights and filter numbers to construct the network. With the block-wise design, the network can not only achieves high performance but also generalizes well to different datasets and tasks. Unlike previous research on automating neural network design which generates the entire network directly, we aim at designing the \texttt{block} structure.

\begin{figure}[tbp]
	\centering
	\includegraphics[width=\linewidth]{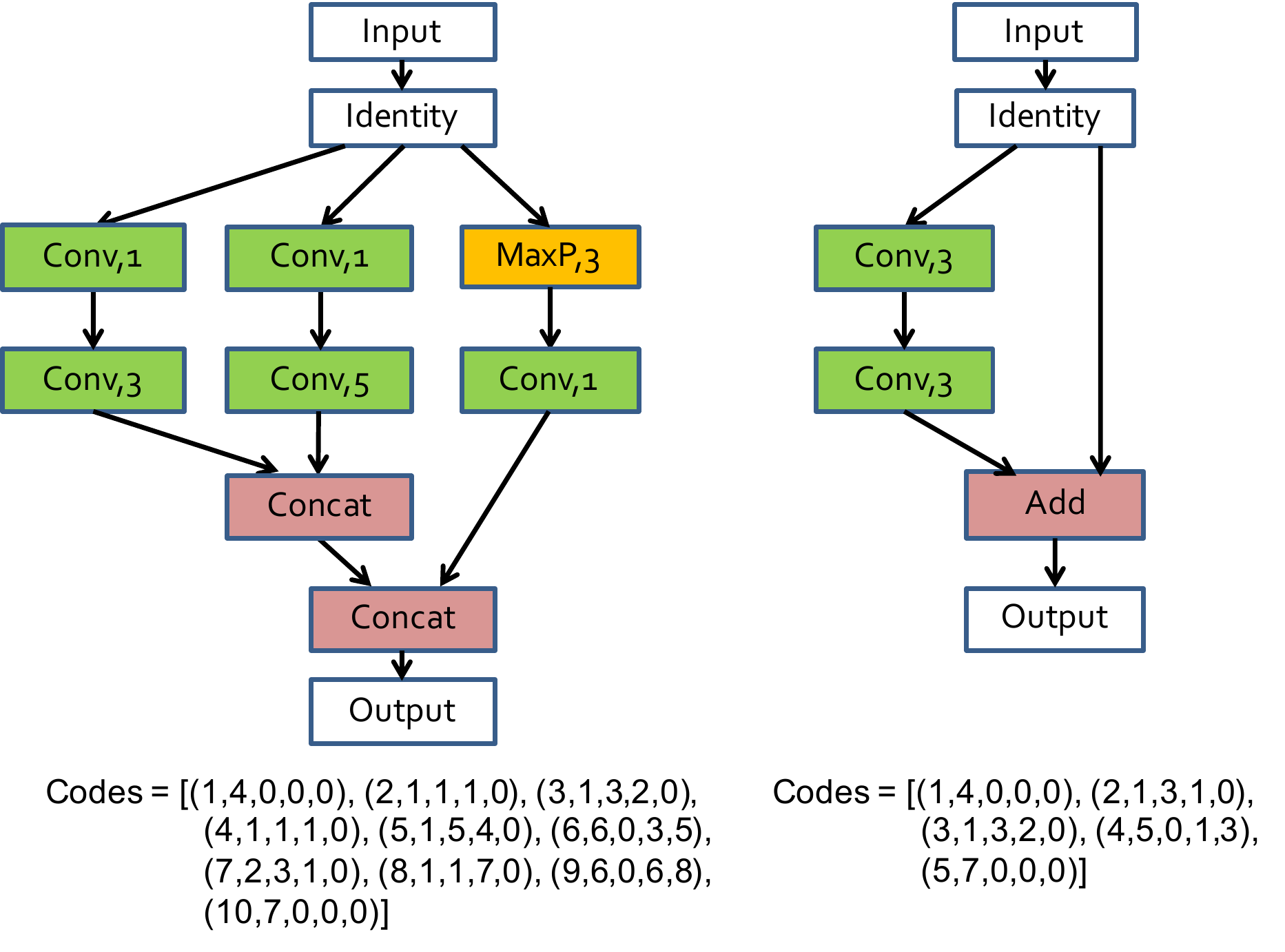}
	\caption{Representative block exemplars with their Network structure codes (NSC) respectively: the block with multi-branch connections (left) and the block with shortcut connections (right).}
	%\vspace{-0.3cm}
	\label{fig:block_codes}
\end{figure}

\begin{figure}[tbp]
	%\linewidth=1
	\centering
	\includegraphics[width=\linewidth]{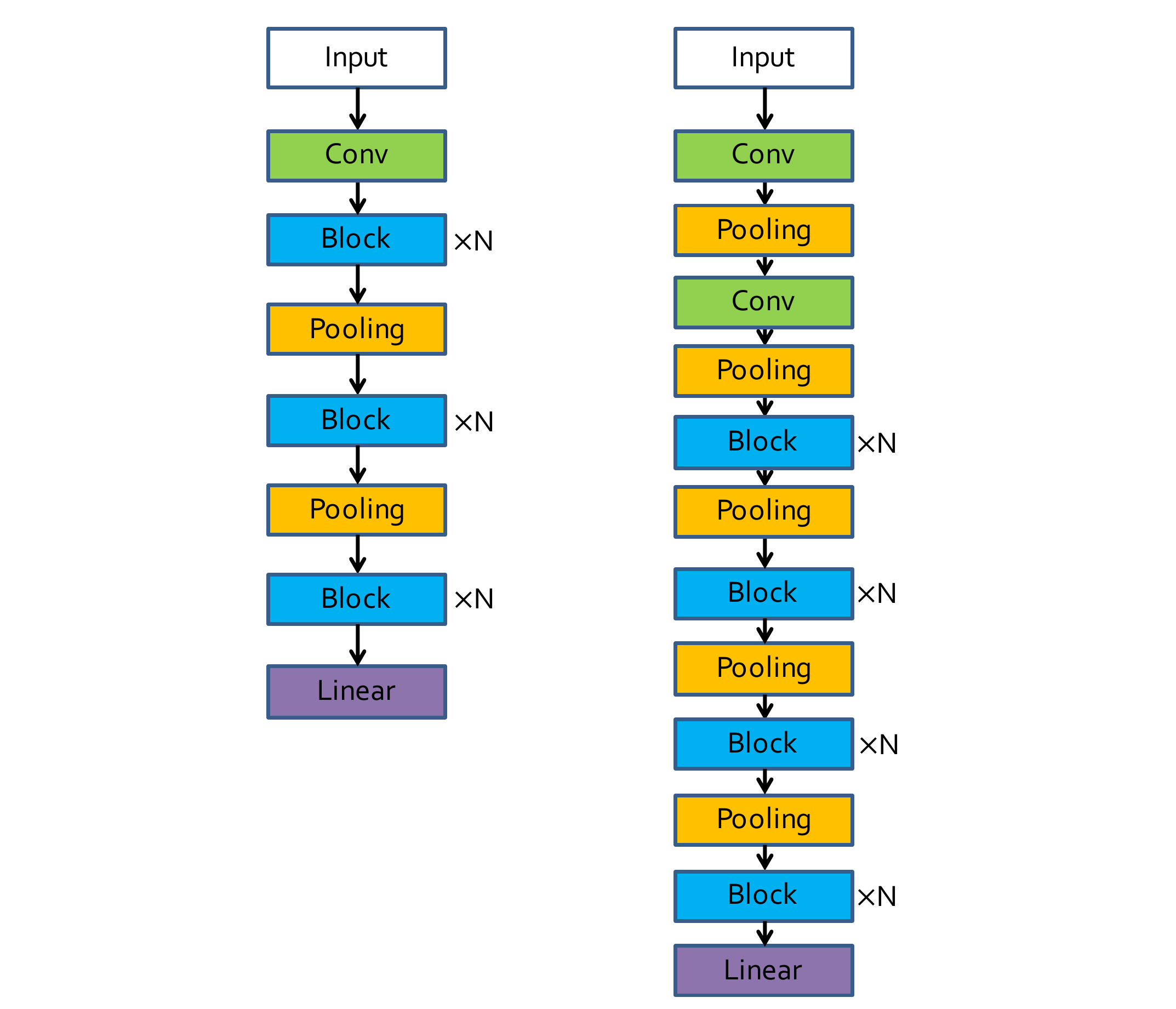}
	\caption{Auto-generated networks on CIFAR-$10$ (left) and ImageNet (right). Each network starts with a few convolution layers to learn low-level features, and followed by multiple repeated \texttt{blocks} with several pooling layers inserted for downsampling.}
	%\vspace{-0.3cm}
	\label{fig:architectures}
\end{figure}

\begin{figure*}[tbp]
	\centering
	\includegraphics[width=\linewidth]{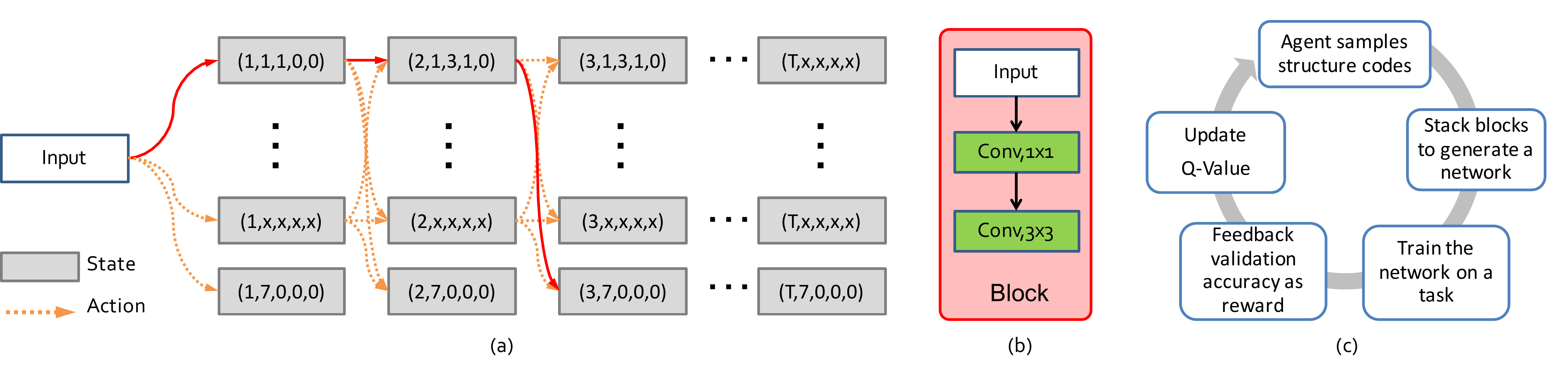}
	\caption{Q-learning process illustration. (a) The state transition process by different action choices. The block structure in (b) is generated by the red solid line in (a). (c) The flow chart of the Q-learning procedure.}
	\label{fig:state_action}
\end{figure*}

As a CNN contains a feed-forward computation procedure, we represent it by a directed acyclic graph (DAG), where each node corresponds to a layer in the CNN while directed edges stand for data flow from one layer to another. To turn such a graph into a uniform representation, we propose a novel layer representation scheme called \textbf{Network Structure Code} (NSC), as shown in Table~\ref{table:1}. Each block is then depicted by a set of $5$-D NSC vectors. In NSC, the first three numbers stand for the \textit{layer index}, \textit{operation type} and \textit{kernel size} respectively. The last two are \textit{predecessor parameters} which refer to the position of a layer's predecessor layer in structure codes. The second predecessor (Pred$2$) is set for the layer owns two predecessors, and for the layer with only one predecessor, Pred$2$ will be set to zero. This design is motivated by the current powerful hand-crafted networks like Inception and Resnet which own their special block structures. This kind of block structure shares similar properties such as containing more complex connections, \eg~shortcut connections or multi-branch connections, unlike the simple connections in plain networks such as AlexNet. Thus, the proposed NSC can encode complexity architectures as shown in Fig.~\ref{fig:block_codes}. 
In addition, all layers without successor in the block are concatenated together to provide the final output.
Note that each convolution operation, same as the declaration in Resnet~\cite{he2016identity}, refers to a \textbf{Pre-activation Convolutional Cell} (PCC) with three components, \ie~\textit{ReLU}, \textit{Convolution} and \textit{Batch Normalization}.
This results in a smaller search space than that with three components separately searchable, and hence with the PCC, we can get better initialization for searching and generating optimal block structure with a quick training process.
% 

% (a) Network architectures we used for classification. They only consist of normal convolution layers, pooling layers and repeated blocks. The top one is for ImageNet task and the bottom one for CIFAR-10. 

Based on the above defined \texttt{blocks}, we construct the complete network by stacking these block structures sequentially which turn a common plain network into its counterpart block version.
Two representative auto-generated networks on CIFAR and ImageNet tasks are shown in Fig.~\ref{fig:architectures}. 
There is no down-sampling operation within each block. We perform down-sampling directly by the pooling layer. If the size of feature map is halved by pooling operation, the block's weights will be doubled.
% All blocks have the same feature map size between the pooling layers, and after pooling layer with a down sampling operation stride of two, feature map will be reduced by two but the block's weights will be doubled. here is no down sampling operation in block.
% 
The architecture for ImageNet contains more pooling layers than that for CIFAR because of their different input sizes, \ie~$224\times 224$ for ImageNet and $32\times 32$ for CIFAR.
More importantly, the \texttt{blocks} can be repeated for arbitrary $N$ times to fulfill different demands, and can even be placed in other manner, such as inserting the block into the Network-in-Network~\cite{lin2013network} framework or setting short cut connection between different blocks.
We will discuss the \texttt{block} connection later in the Section~\ref{subsec:block_conn}.

\subsection{Designing Network Blocks With Q-Learning}
\label{subsec:q_learning}

Albeit we squeeze the search space of the entire network design by focusing on constructing network \texttt{blocks}, there is still a large amount of possible structures to seek. Therefore, we employ reinforcement learning rather than random sampling for automatic design. Our method is based on standard tabular Q-learning, a kind of reinforcement learning, which concerns how an agent ought to take actions so as to maximize the cumulative reward. The Q-learning model consists of an \textit{agent}, \textit{states} and a set of \textit{actions}. 

In this paper, the \textit{state} $s\in S$ represents the status of the current layer which is defined as a Network Structure Code (NSC) claimed in Section~\ref{subsec:blocks}, \ie~$5$-D vector \{layer index, layer type, kernel size, pred$1$, pred$2$\}. The \textit{action} $a\in A$ is the decision for the next successive layer. 
Thanks to the defined NSC set with a limited number of choices, both the \textit{state} and \textit{action} space are thus finite and discrete to ensure a relatively small search space. 
The state transition process $(s_t,a(s_t))\rightarrow (s_{t+1})$ is shown in Fig.~\ref{fig:state_action}(a), where $t$ refers to the current layer. The block example in Fig.~\ref{fig:state_action}(b) is generated by the red solid lines in Fig.~\ref{fig:state_action}(a).
% 
% In addition, we restrict the environment to have a discrete and finite state space \(S\) as well as action space \(A\) to ensure a relatively small search space. Figure. \ref{fig:state_action}(a) shows the state \(s \in S\), the action \(a \in A\)  and state transition process\((s_t,A(s_t)) \rightarrow (s_{t+1})\). 
% 
The learning agent is given the task of sequentially picking NSC of a block. The structure of block can be considered as an action selection trajectory $\tau_{a_{1:T}}$, \ie~a sequence of NSCs. 
We model the layer selection process as a Markov Decision Process with the assumption that a well-performing layer in one block should also perform well in another block~\cite{baker2016designing}.
To find the optimal architecture, the agent maximizes its expected \textit{reward} over all possible trajectories, denoted by \(R_{\tau}\),
\begin{eqnarray}
R_{\tau}= \mathbb{E}_{P(\tau_{a_{1:T}})}[\mathbb{R}],
\end{eqnarray}
where the \(\mathbb{R}\) is the cumulative reward. The expected reward can be maximized using the recursive Bellman Equation. Given a state \(s_t\in S\) and subsequent action \(a\in A(s_t)\), we define the maximum total expected reward to be \(Q^*(s_t,a)\) which is known as Q-value of state-action pair. The recursive Bellman Equation is then written as 
\begin{eqnarray}
\nonumber Q^*(s_t,a)=\mathbb{E}_{s_{t+1}|s_t,a}[\mathbb{E}_{r|s_t,a,s_{t+1}}[r|s_t,a,s_{t+1}]\\
+\gamma \max_{a'\in A(s_{t+1}))}Q^*(s_{t+1},a')].
\end{eqnarray}

%An empirical iterative approximation of the above quantity is
%it can be formulated as an iterative update
Empirically, the above quantity can be formulated as an iterative update:
\begin{align}
Q(s_T,a) =& 0,\\
Q(s_{T-1},a_T) =& (1-\alpha)Q(s_{T-1},a_T) + \alpha r_T,\\
\nonumber Q(s_t,a)=&(1-\alpha)Q(s_t,a)\\
+ \alpha [r_t+\gamma \max_{a'}&Q(s_{t+1},a')], t \in\{1,2,...T-2\},
\end{align}
%\begin{multline}
%Q_{i+1}(s_t,a)=Q_{i}(s_t,a) +\\
%\alpha [r_t+\gamma \max_{a'\in A(s_{t+1}))}Q_i(s_{t+1},a')-Q_{i}(s_t,a)],
%\end{multline}
%for example, we can design a function to describe every layer's contribution for the performance of block.
where \(\alpha\) is the learning rate which determines how the newly acquired information overrides the old information, \(\gamma\) is the discount factor which measures the importance of future rewards;
%$r_T$ is the validation accuracy of corresponding network trained convergence on training set for final state $s_{T}$, \ie~terminal layers.
\(r_t\) denotes the intermediate reward observed for the current state \(s_{t}\), and $s_{T}$ refers to final state,~\ie terminal layers; $r_T$ is the validation accuracy of corresponding network trained convergence on training set for $a_T$,~\ie action to final state. Since the reward \(r_t\) cannot be explicitly measured in our task, we use reward shaping~\cite{ng1999policy} to speed up training. The shaped intermediate reward is defined as:
%we use the validation accuracy with the hyperparameter $\lambda$ as an alternative:
% and the Q-value is reformulated as
%\begin{multline}
%Q_{i+1}(s_t,a) = Q_{i}(s_t,a) +\\
%\alpha[\text{accuracy} +\gamma \max_{{a}'\in A(s_{t+1})}Q_i(s_{t+1},{a}')-Q_{i}(s_t,a)].
%\end{multline}
\begin{eqnarray}
r_t = \frac{r_T}{T}.
\end{eqnarray}

\begin{figure}[tbp]
	\centering
	\includegraphics[width=\linewidth]{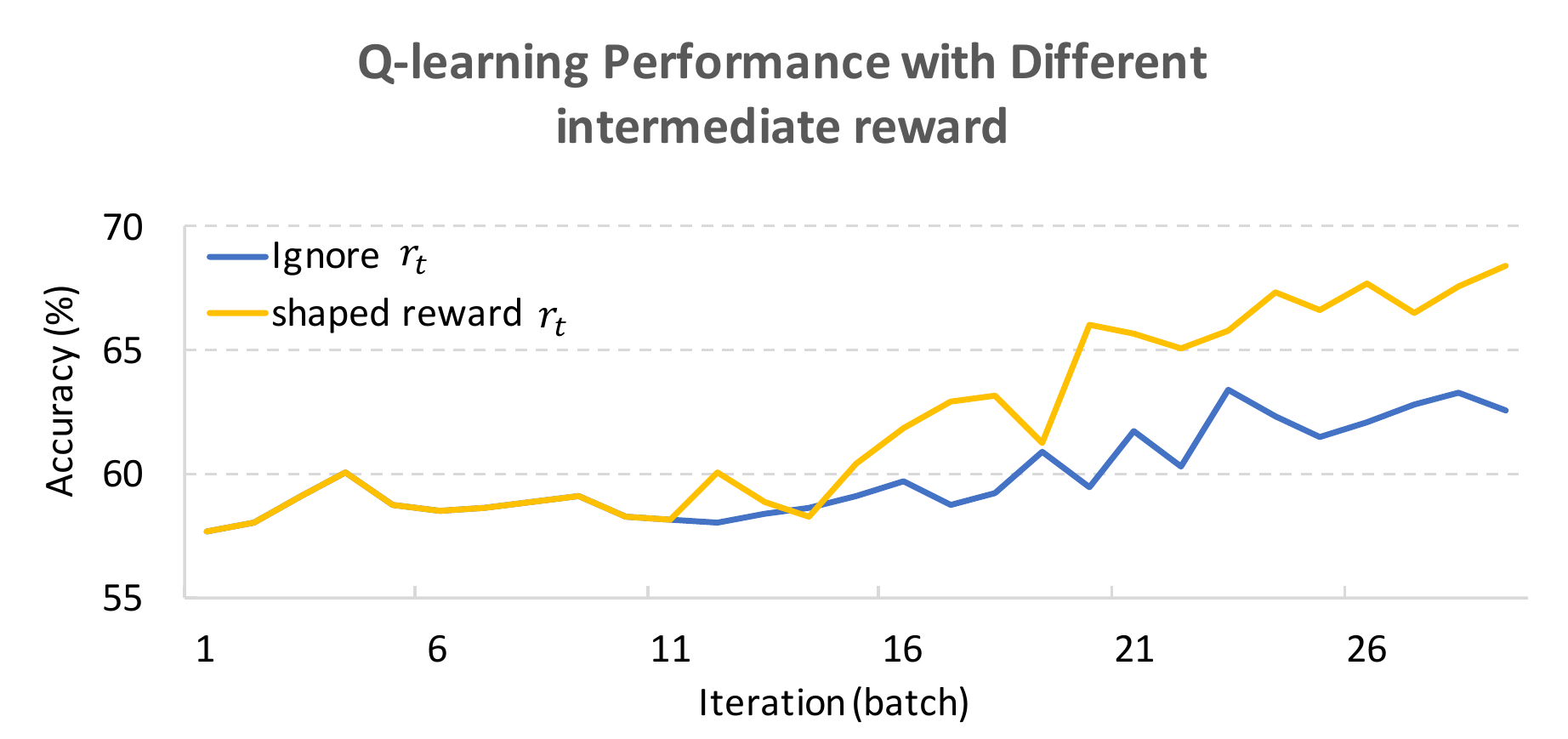}
	\caption{Comparison results of Q-learning with and without the shaped intermediate reward $r_t$. By taking our shaped reward, the learning process convergent faster than that without shaped reward start from the same exploration.}
	\label{fig:acc_reward}
\end{figure}

Previous works~\cite{baker2016designing} ignore these rewards in the iterative process by simply setting them to zero, which may cause a slow convergence in the beginning. This is known as the temporal credit assignment problem which makes RL time consuming~\cite{sutton1998reinforcement}. In this case, the Q-value of $s_T$ is much higher than others in early stage of training and thus leads the agent prefer to stop searching at the very beginning, \ie~tend to build small block with fewer layers.
%\begin{eqnarray}
%Q(s_t,a)\sim \alpha^{f-t}R\ll Q(s_T,a),  \alpha=0.01
%\end{eqnarray}
As the comparison result in Fig.~\ref{fig:acc_reward} shows, the learning process of the agent with our shaped reward \(r_t\) is convergent much faster than previous method.

% Q-learning result with different version of  Bellman¡¯s Equation approximation. The learning process with the \(r_t\), we use test accuracy to represent, can convergence faster than ignore it.

We summarize the learning procedure in Fig.~\ref{fig:state_action}(c). The agent first samples a set of structure codes to build the block architecture, based on which the entire network is constructed by stacking these blocks sequentially. We then train the generated network on a certain task, and the validation accuracy is regarded as the reward to update the Q-value. Afterwards, the agent picks another set of structure codes to get a better block structure.

\subsection{Early Stop Strategy}
\label{subsec:early_stop}

\begin{figure}[tbp]
	\centering
	\includegraphics[width=\linewidth]{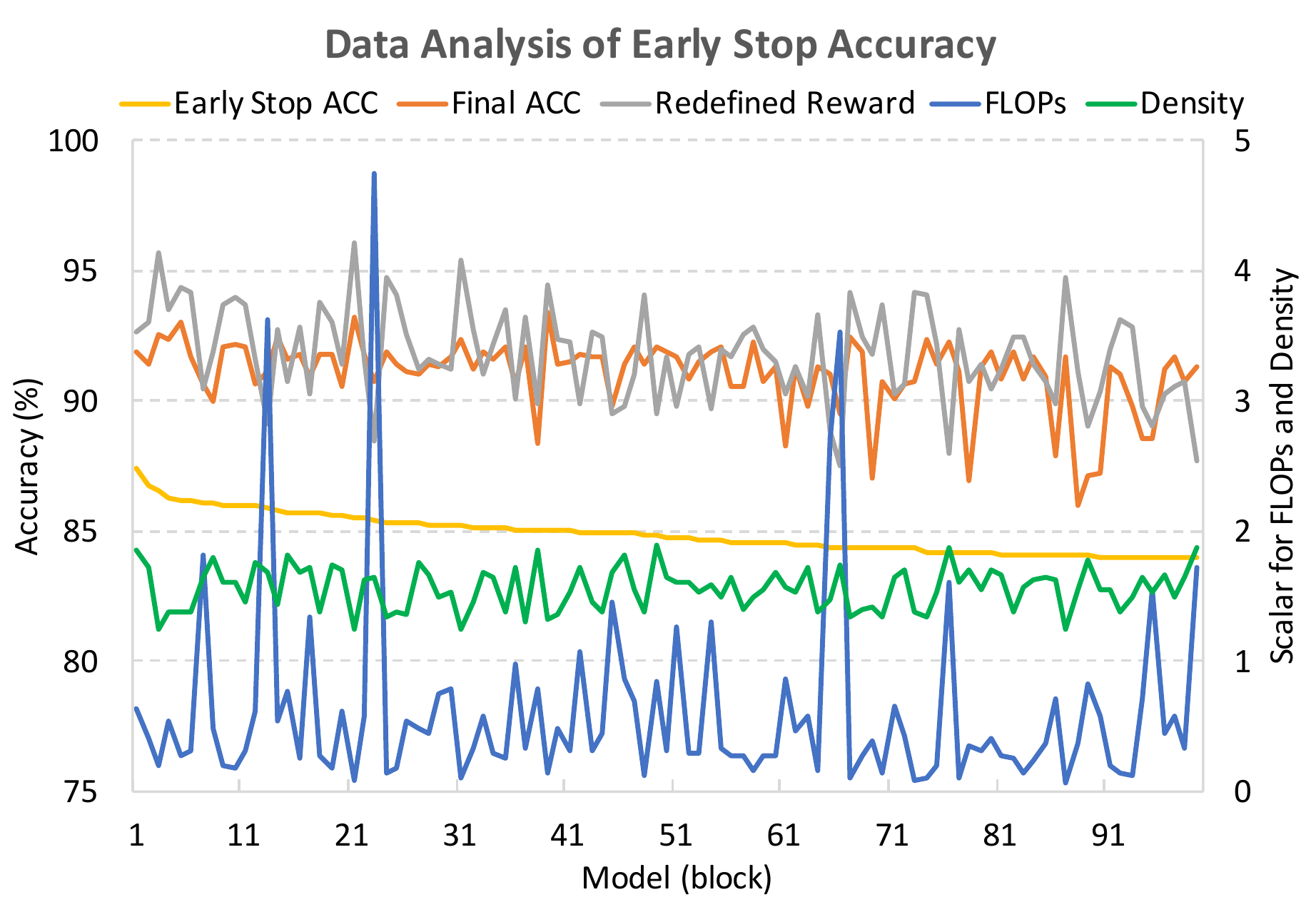}
	\caption{The performance of early stop training is poorer than the final accuracy of a complete training. With the help of FLOPs and Density, it squeezes the gap between the redefined reward function and the final accuracy.}
	\label{fig:early_stop}
\end{figure}

Introducing \texttt{block}-wise generation indeed increases the efficiency. However, it is still time consuming to complete the search process. To further accelerate the learning process, we introduce an early stop strategy. However, early stopping training process may result in a poor accuracy. Fig.~\ref{fig:early_stop} shows an example, where the early-stop accuracy in yellow line is much lower than the final accuracy in orange line, which means that some good blocks unfortunately perform worse than bad blocks when stop training early. Meanwhile, we notice that the FLOPs and density of the corresponding blocks have a negative correlation with the final accuracy. Thus, we redefine the reward function as 
\begin{eqnarray}
\nonumber reward = \text{ACC}_{\text{EarlyStop}} - \mu\log(\text{FLOPs})\\
- \rho\log(\text{Density}),
\label{eq:early_stop_reward}
\end{eqnarray}
where FLOPs~\cite{he2015convolutional} refer to an estimation of computational complexity of the block, and Density is the edge number divided by the dot number in DAG of the block. There are two hyperparameters, $\mu$ and $\rho$, to balance the weights of FLOPs and Density. With the redefined reward function, the reward is more relevant to the final accuracy.

With this early stop strategy and small search space of network blocks, the training process just costs $3$ days to complete the searching process with only $32$ GPUs, which is superior to that of~\cite{zoph2016neural}, which spends $28$ days with $800$ GPUs to achieve the similar performance.
However, the use of 32 GPUs is still not common for most deep learning practitioners.
Accordingly, we will further accelerate the searching process using an algorithm called faster BlockQNN described in Section~\ref{subsec:predict_network_prf}.

\subsection{Connection Style Between Blocks}
\label{subsec:block_conn}

Our method can search for the optimal block structure effectively and efficiently, but still utilizes manual rules, \ie~stacking the block structures sequentially and when the size of feature map is halved by pooling operation, the block's weights will be doubled or increased. With this block connection rule, we can transfer the block structure between different datasets and tasks easily.

However, stacking the block structures sequentially may not be the optimal connection style between blocks. Hence, we trade off some transferable ability and design the connection between specific block structures automatically by search. The only difference from the above method lies in the definition of Network Structure Code: \textit{convolution layer} is replaced by \texttt{blocks structure} and the \textit{kernel size} for convolutional operation is substituted by \textit{channel numbers}. Firstly, the block structures are connected sequentially and we use the \textit{predecessor parameters} to represent the additional connection between different blocks. We perform down-sampling only by the pooling layer with stride $2$ and use$1\times1$ convolutions to match the different dimensions for connected layers. 

With the block connection auto-generation module, the BlockQNN can be regarded as a two-stage framework: (1) find the optimal block, and (2) find the optimal connection for optimal block. We can further improve the performance of auto-generated network with this two-stage strategy, it proves stacking the block structures sequentially is not the best choice. But the generated network is dataset-specific that cannot be well transferred to other tasks with different input data size. The exploration of universal block connection formulation with transferable ability is still an open problem.

\subsection{Predicting Network Performance Before Training}
\label{subsec:predict_network_prf}

\begin{figure*}[t]
	\begin{center}
		\includegraphics[width=0.95\linewidth]{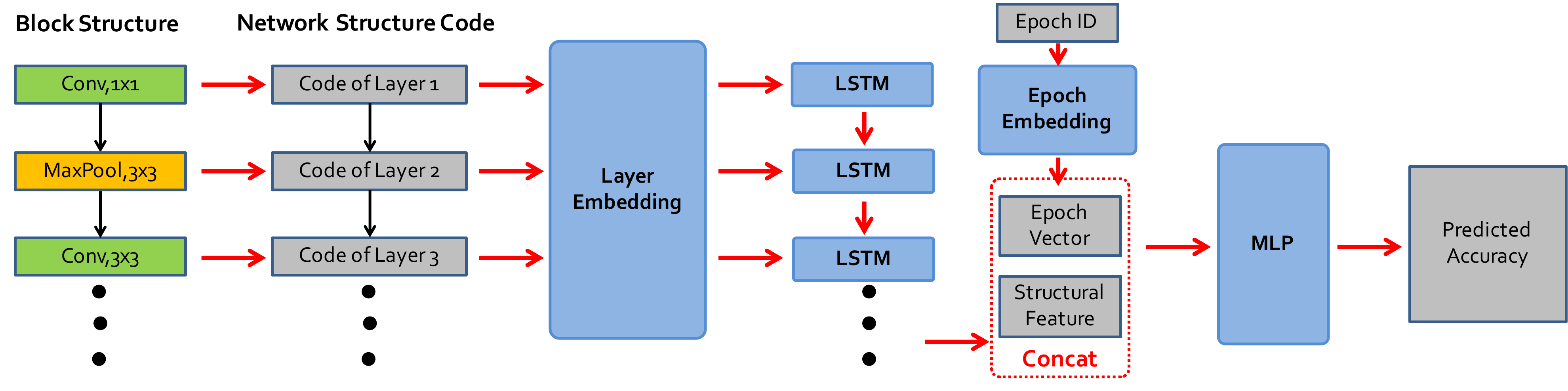}
	\end{center}
	\caption{\small
		The overall pipeline of the \emph{Faster BlockQNN} framework.
		Given a network architecture, it first encodes each layer into
		a vector through integer coding and layer embedding.
		Subsequently, it applies a recurrent network with LSTM units to
		integrate the information of individual layers following the
		network topology into a \emph{structural feature}.
		This structural feature together with the epoch index (also embedded
		into a vector) will finally be fed to an MLP to predict
		the accuracy at the corresponding time point, \ie~the end of the
		given epoch.
		Note that the blocks indicated by blue color,
		including the embeddings, the LSTM, and the MLP,
		are jointly learned in an end-to-end manner.
	}
	\label{fig:overview}
\end{figure*}

%% Problem statement
%Although we can complete the searching process on CIFAR use only $32$ GPUs in 3 days with small search space of network blocks and early stop strategy, the hardware requirement of $32$ GPUs is not so easy to satisfy for the common deep learning practitioner.
To further accelerate the block searching process, for the common deep learning practitioners who have limited computing resources, we propose a strategy to predict the network performance before training.
As we know, the most time-consuming portion in network generation is the training of the sampled network to get the validation accuracy as reward. To mitigate this cost, we assess a network architecture quantitatively before investing resources in training it.
The design method with network performance prediction is called as Faster BlockQNN, which is depicted in Figure~\ref{fig:overview}.

The network performance prediction model can be formalized as a function, denoted by $f$.
The function $f$ takes two arguments,
a network architecture $x$ and an epoch index $t$, and produces
a scalar value $f(x, t)$ as the prediction of the accuracy
at the end of the $t$-th epoch.
Here, incorporating the epoch index $t$ as an input to $f$ is reasonable,
as the validation accuracy generally changes as the training proceeds.
Therefore, when we predict performance, we have to be specific about
the time point of the prediction.

%% Differences from previous formulations

Note that this formulation differs fundamentally from
previous works~\cite{domhan2015speeding,klein2016learning,baker2017practical}, which require the observation of the initial part (usually $25\%$)
of the training curve and extrapolate the remaining part.
On the contrary, our method aims to predict the entire curve, relying only
on the network architecture.
In this way, it can provide feedback much quicker and thus is particularly
suited for large-scale search of network designs.

%% Section content

%However, developing such a predictor is nontrivial.
%Towards this goal, we are facing significant technical challenges,
%\eg~\emph{how to unify the representation of various layers},
%and \emph{how to integrate the information from individual layers
%	over various network topologies.}
%%
%In what follows, we will present our answers to these questions.
%Particularly, Sec.~\ref{subsec:layercode} presents a unified vector representation of layers, 
%which is constructed in two steps, namely coding and embedding.
%Sec.~\ref{subsec:model} presents an LSTM model for integrating the
%information across layers.

%% Overview of ULC

\label{subsec:embedding}
\begin{figure}[t]
	\begin{center}
		\includegraphics[width=0.95\linewidth]{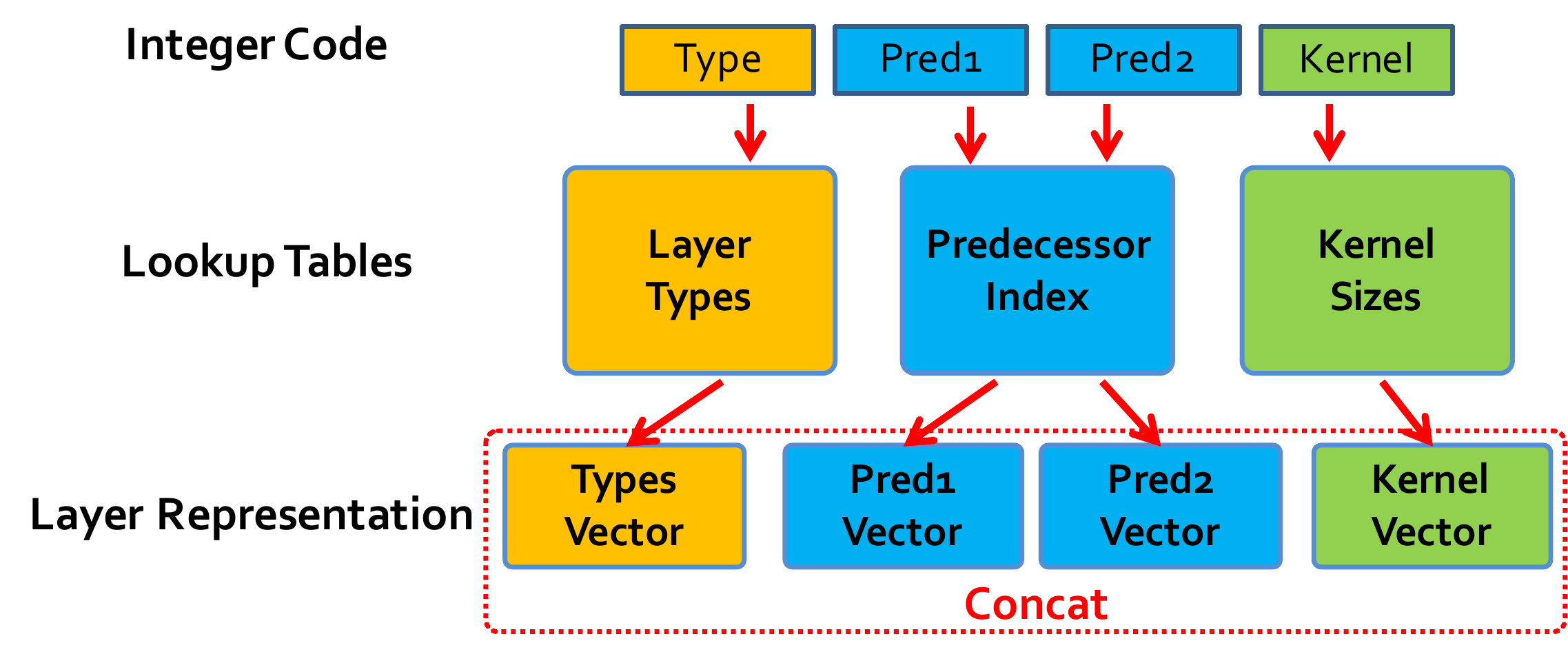}
	\end{center}
	\caption{\small
		The layer embedding component. It takes the integer codes
		as input, maps them to embedded vectors respectively via
		table lookup, and finally concatenates them into a real
		vector representation. Note that \emph{Pred1} and \emph{Pred2}
		share the same lookup table.
	}
	\label{fig:embedding}
\end{figure}

%Layer embedding.
%

As stated in Section~\ref{subsec:blocks}, we define the various layers in network as Network Structure Code (NSC), \ie~$5$-D vector \{layer index, layer type, kernel size, pred$1$, pred$2$\}.
While capturing the key information for a layer,
this \emph{discrete} representation is not amenable to
complex numerical computation and deep pattern recognition.
Inspired by word embedding~\cite{mikolov2013distributed},
a strategy proven to be very effective in natural language processing,
we take one step further and develop
\emph{Layer Embedding}, a scheme to turn the integer codes into
a unified real-vector representation.

As shown in Figure~\ref{fig:embedding}, the embedding is done by table
lookup. Specifically, this module is associated with three lookup tables,
respectively for \emph{layer types}, \emph{kernel sizes}, and
\emph{predecessor}. Note that the predecessor table is used to encode
both \emph{Pre1} and \emph{Pre2}.
Given a tuple of integers, we can convert its element into a real vector
by retrieving from the corresponding lookup table.
Then by concatenating all the embedded vectors derived respectively
from individual integers, we can form a vector representation of the layer.

%Integrated Prediction
With the layer-wise representations based on Network Structure Code (NSC) and Layer
Embedding, the next is to aggregate them into an overall representation
for the entire network.
Inspired by the success of recurrent networks in sequential modeling,
\eg~in language modeling~\cite{mikolov2010recurrent} and video analytics~\cite{yue2015beyond},
we choose to explore recurrent networks in our problem.
Specifically, we adopt the \emph{Long-Short Term Memory (LSTM)}~\cite{hochreiter1997long},
an effective variant of RNN, for integrating the information along
a sequence of layers.
Along the way from low-level to high-level layers, the LSTM network
would gradually incorporate layer-wise information into the hidden state.
At the last step, \ie~the layer right before the fully connected layer
for classification, we extract the hidden state of the LSTM cell to represent
the overall structure of the network, which we refer to as the
\emph{structural feature}. 
% For Reversed Sequence
Note that we would reverse the topological sequence of architectures before feeding it to LSTM since we think Long-term dependencies are better recognized by LSTM if the signal occurs at the starting point.

As shown in Figure~\ref{fig:overview}, the \emph{Faster BlockQNN} framework will
finally combine this structural feature with the epoch index (also embedded
into a real-vector) and use a Multi-Layer Perceptron (MLP) to make the final
prediction of accuracy.
In particular, the MLP component at the final step is comprised of
three fully connected layers with Batch Normalization and ReLU activation.
The output of this component is a real value that serves as an estimate
of the accuracy.

%Learning Objective
Given a set of sample networks $\{x_i\}_{1:N}$,
we can obtain a performance curves $y_i(t)$ for each network $x_i$,
\ie~the validation accuracy as a function of epoch numbers,
by training the network on a given dataset.
Hence, we can obtain a set of pairs $\mathcal{D} = \{(x_i, y_i)\}_{1:N}$
and learn the parameters of the predictor in a supervised way.

Specifically, we formulate the learning objective with the \emph{smooth L1} loss,
denoted by $l$, as below:
\begin{equation}
\mathcal{L}(\mathcal{D}; \boldsymbol\theta) =
\frac{1}{N} \sum_{i=1}^n l \left(f(x_i, T), y_i(T)\right).
\end{equation}
Here, $\boldsymbol\theta$ denotes the predictor parameters.
Note that we train each sample network with $T$ epochs, and use the results of
the final epoch to supervise the learning process.
Our framework is very flexible -- with the entire learning curves, in principle,
one can use the results at multiple epochs for training.
However, we found empirically that using only the final epochs already yields
reasonably good results.

With the Faster BlockQNN, the demand of computing resource will be further reduced.
More accurately, we can get a comparable result with only 1 GPU in 20 hours which is afforded for common deep learning researcher.

\section{ Training Details}

%\subsection{Distributed Asynchronous Framework}

\begin{figure}[tbp]
	\centering
	\includegraphics[width=\linewidth]{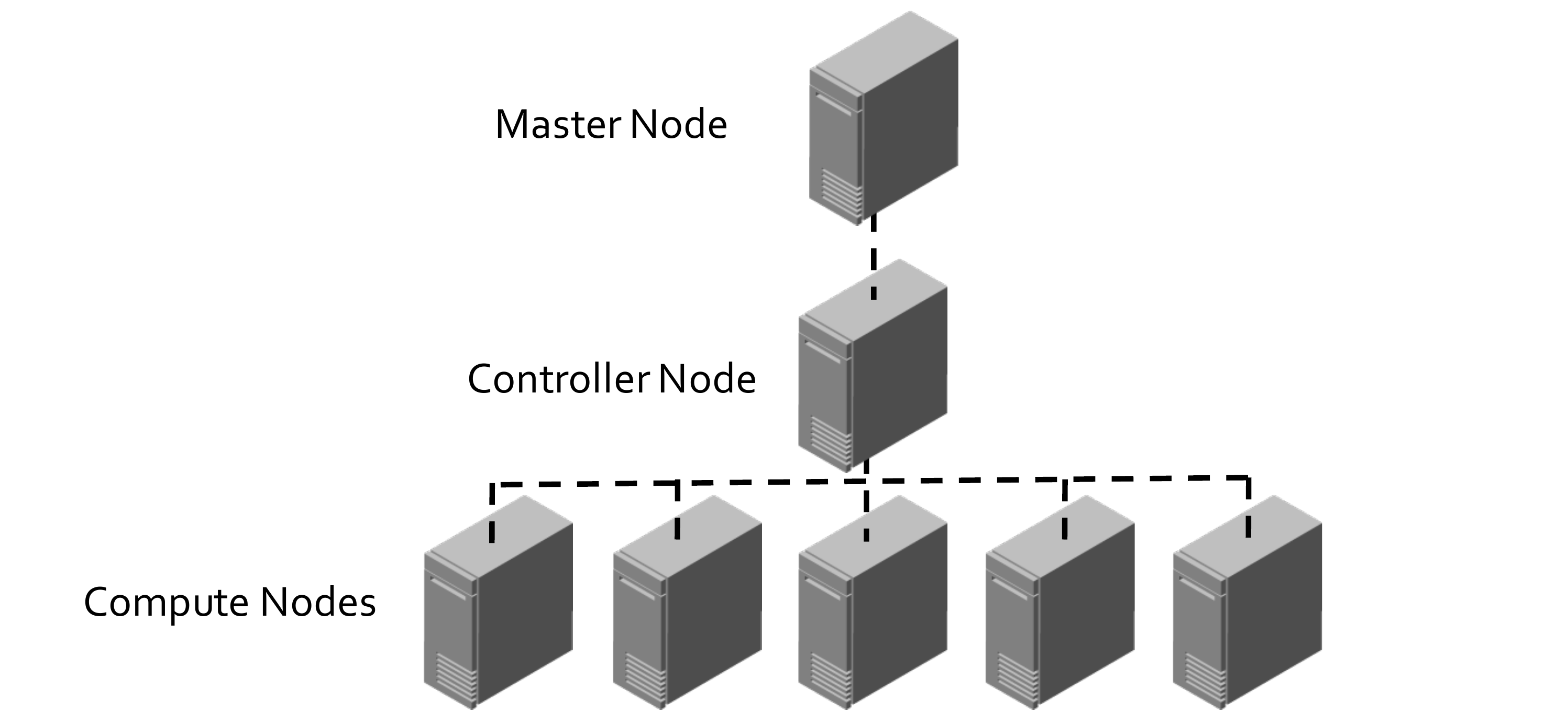}
	\caption{The distributed asynchronous  framework. It contains three parts: master node, controller node and compute nodes.}\label{fig:ps}
\end{figure}

\vspace{0.1cm}\noindent \textbf{Distributed Asynchronous Framework.} To speed up the learning of agent, we use a distributed asynchronous framework as illustrated in Fig.~\ref{fig:ps}. It consists of three parts: master node, controller node and compute nodes. The agent first samples a batch of block structures in master node. Afterwards, we store them in a controller node which uses the block structures to build the entire networks and allocates these networks to compute nodes. 
It can be regarded as a simplified parameter-server~\cite{dean2012large,li2013parameter}. Specifically, the network is trained in parallel on each of compute nodes and returns the validation accuracy as reward by controller nodes to update agent. With this, we can generate network efficiently on multiple machines with multiple GPUs.

%\subsection{Training Details}

\vspace{0.1cm}\noindent \textbf{Epsilon-greedy Strategy.} The agent is trained using Q-learning with experience replay~\cite{lin1993reinforcement} and epsilon-greedy strategy~\cite{mnih2015human}. %Table~\ref{table:2} shows the epsilon schedule we used in the experiments. 
With epsilon-greedy strategy, the random action is taken with probability $\epsilon$ and the greedy action is chosen with probability $1-\epsilon$. We decrease epsilon from $1.0$ to $0.1$ following the epsilon schedule as shown in Table~\ref{table:2} such that the agent can transform smoothly from exploration to exploitation. We find that the result goes better with a longer exploration, since the search scope would become larger and the agent can see more block structures in the random exploration period. 
%
%Table~\ref{table:2} shows that the number of batches for different epsilon we used in the experiments. We decrease epsilon from 1.0 to 0.1, leaving half the searching time for exploration.
%

\vspace{0.1cm}\noindent\textbf{Experience Replay.} Following~\cite{baker2016designing}, we employ a replay memory to store the validation accuracy and block description after each iteration. Within a given interval, \ie~each training iteration, the agent samples $64$ blocks with their corresponding validation accuracies from the memory and updates Q-value $64$ times.
%table 2
\begin{table}[t!]
	%% increase table row spacing, adjust to taste
	\renewcommand{\arraystretch}{1.3}
	%\centering

	\begin{center}
    	\caption{Epsilon Schedules. The number of iteration the agent trains at each epsilon($\epsilon$) state.}\label{table:2}			
		\scriptsize
		\begin{tabular}{c|c|c|c|c|c|c|c|c|c|c}
			\hline
			$\epsilon$&1.0& 0.9 & 0.8 & 0.7 & 0.6 & 0.5 & 0.4 & 0.3 & 0.2 & 0.1 \\
			\hline
			Iters&95& 7 & 7 & 7 & 10 & 10 & 10 & 10 & 10 & 12 \\
			\hline
		\end{tabular}
	\end{center}

\end{table}

\vspace{0.1cm}\noindent \textbf{BlockQNN Generation.} In the Q-learning update process, the learning rate \(\alpha\) is set to 0.01 and the discount factor \(\gamma\) is $1$. We set the hyperparameters $\mu$ and $\rho$ in the redefined reward function as $1$ and $8$, respectively. The agent samples $64$ sets of NSC vectors at a time to compose a mini-batch and the maximum layer index for a block is set to $23$. We train the agent with $178$ iterations, \ie~sampling $11,392$ blocks in total. 

During the block searching phase, the compute nodes train each generated network for a fixed $12$ epochs on CIFAR-$100$ using the early top strategy as described in Section~\ref{subsec:early_stop}. CIFAR-100 contains $60,000$ samples with $100$ classes which are divided into training and test set with the ratio of $5:1$. We train the network without any data augmentation procedure. 
The batch size is set to 256. We use Adam optimizer~\cite{kingma2014adam} with \(\beta_1 = 0.9  \), \(\beta_2 = 0.999  \), \(\varepsilon =10^{-8} \). The initial learning rate is set to 0.001 and is reduced with a factor of $0.2$ every $5$ epochs. All weights are initialized as in~\cite{he2015delving}. 
If the training result after the first epoch is worse than the random guess, we reduce the learning rate by a factor of $0.4$ and restart training, with a maximum of $3$ times for restart-operations. 

After obtaining one optimal block structure, we build the whole network with stacked blocks and train the network until converging to get the validation accuracy as the criterion to pick the best network. In this phase, we augment data with randomly cropping the images with size of $32\times 32$ and horizontal flipping. Besides, we also apply the cutout regularization during training~\cite{devries2017cutout}. All models use the SGD optimizer with momentum rate set to $0.9$ and weight decay set to $0.0005$. We start with a learning rate of $0.1$ and train the models for $300$ epochs with a single period cosine decay as in~\cite{loshchilov2016sgdr}. The batch size is set to $128$ and all weights are initialized with MSRA initialization~\cite{he2015delving}.

\vspace{0.1cm}\noindent \textbf{Transferable BlockQNN.} 
We also evaluate the transferability of the best auto-generated block structure searched on CIFAR-$100$ to a smaller dataset, CIFAR-$10$, with only $10$ classes and a larger dataset, ImageNet, containing $1.2$M images with $1000$ classes. All the experimental settings on CIFAR-$10$ are the same as those on the CIFAR-$100$ stated above.
ImageNet models are trained on $224$x$224$ images and evaluated on $224$x$224$ or $320$x$320$ images with center crop. The training is conducted with a mini-batch size of $512$ where each image has the same data augmentation procedures as described previously~\cite{szegedy2015rethinking}, and is optimized with SGD strategy. The initial learning rate, weight decay and momentum are set as $0.1$, $0.0001$ and $0.9$, respectively. We decay the learning rate with a single period cosine annealing as in~\cite{loshchilov2016sgdr}.
Additionally, we use label smoothing with a value of $0.1$ and an auxiliary classifier located at $2/3$ of the way up the network with the weight of $0.4$ for all ImageNet models as done in~\cite{szegedy2015rethinking}. Dropout is applied to the final softmax with probability $0.4$.

\begin{figure*}[tbp]
	\centering
	\includegraphics[width=\linewidth]{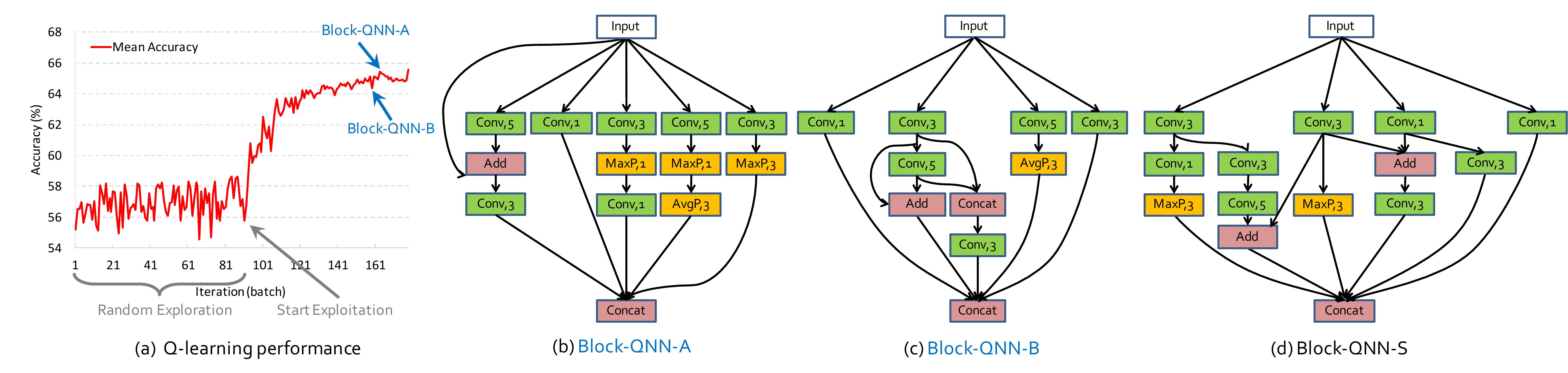}
	%\vspace{0.05cm}
	\caption{(a) Q-learning performance on CIFAR-$100$. The accuracy goes up with the epsilon decrease and the top models are all found in the final stage, show that our agent can learn to generate better block structures instead of random searching. (b-c) Topology of the Top-$2$ block structures generated by our approach. We call them Block-QNN-A and Block-QNN-B. (d) Topology of the best block structures generated with limited parameters, named Block-QNN-S.  }
	\label{fig:8}
\end{figure*}

\vspace{0.1cm}\noindent \textbf{Block Connection.} 
The training process is basically same as the block searching phase. The agent samples $64$ sets of NSC vectors for Block Connection at a time to compose a mini-batch and the maximum layer index for a network is set to $12$. We train the agent with $46$ iterations, \ie~sampling $2,944$ connection styles in total. Specifically, we allow $5$ pooling layers at most in CIFAR task to ensure the resolution is not too small. Moreover, we use the Block-QNN-S, introduced in section~\ref{sec:result_cifar}, as the basic block structure for the connection generation.

\vspace{0.1cm}\noindent \textbf{Faster BlockQNN.} 
For the Faster BlockQNN model, we use $40$-dimensional vectors for both layer embedding and epoch embedding. The dimension of the hidden states in LSTM is set to $160$. The Multi-Layer Perceptron (MLP) for final prediction comprises $3$ linear layers, each with $200$ hidden units. We randomly sample $2000$ block structures and train them on CIFAR-$100$ to get performance curves for the predictor, the experimental setup is same as the block searching phase. After that, the training process for sampled block structures is replaced by the Faster BlockQNN predictor in the searching phase, other setting is the same as the standard BlockQNN generation. 
% 

%\color{cyan} [not necessary?? if space is not enough, suggest remove this paragraph.] 
Our framework is implemented under the PyTorch scientific computing platform. We use the CUDA backend and cuDNN accelerated library in our implementation for high-performance GPU acceleration. Our experiments are carried out on $32$ NVIDIA $1080$Ti GPUs and took about $3$ days to complete searching.
Moreover, the faster version can also get a comparable result with only 1 GPU in 20 hours.

\begin{figure}[tbp]
	\centering
	\includegraphics[width=\linewidth]{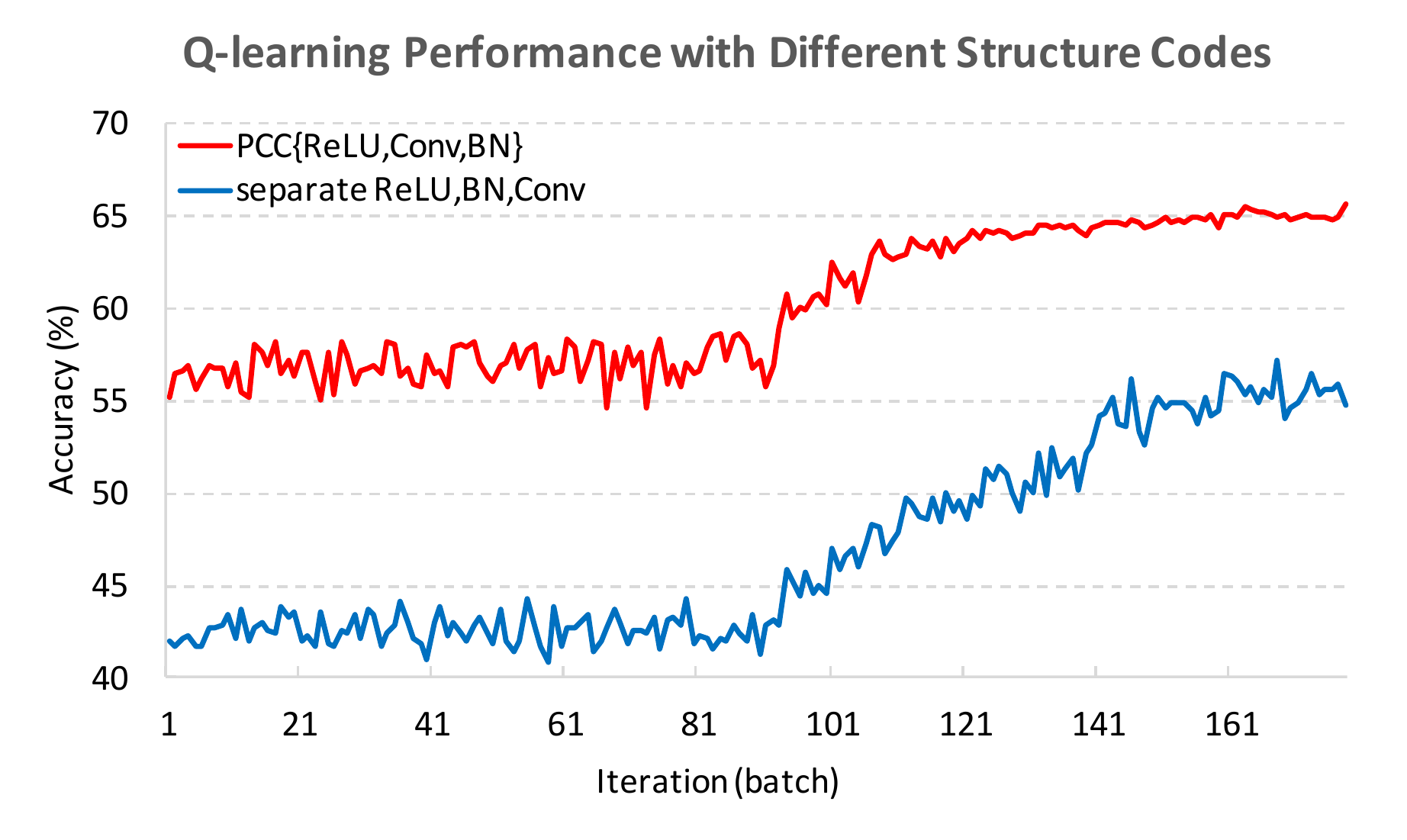}
	\caption{Q-learning result with different NSC on CIFAR-$100$. The red line refers to searching with PCC, \ie~combination of ReLU, Conv and BN. The blue stands for separate searching with ReLU, BN and Conv. The red line is better than blue from the beginning with a big gap.}\label{fig:pcc}
	%\vspace{-0.3cm}
\end{figure}

\section{Experimental Results}
%introductory paragraph
%In this section, we will show the generated characters visually, and analyze the quality of the generated characters by feeding them into the discriminative RNN model to check whether they are recognizable or not. Moreover, we will also discuss properties of the character embedding matrix.
%We evaluate the proposed models on four challenging datasets: PASCAL VOC 2012, PASCAL-Context, PASCAL- Person-Part, and Cityscapes. We first report the main results of our conference version [38] on PASCAL VOC 2012, and move forward to latest results on all datasets.

In this section, we will analyze the training process of block searching and block connection searching. And then, we present experiments on CIFAR and ImageNet, for the purpose of evaluating and comparing the proposed model with other state-of-the-art approaches. Moreover, we will also discuss properties of the evolutionary process and the efficiency of the network architecture generation.

\subsection{Block Searching Analysis}
\label{subsec:block_analysis}
Fig.~\ref{fig:8}(a) provides early stop accuracies over $178$ batches on CIFAR-$100$, each of which is averaged over $64$ auto-generated block-wise network candidates within in each mini-batch. After random exploration, the early stop accuracy grows steadily till converges. The mean accuracy within the period of random exploration is $56\%$ while finally achieves $65\%$ in the last stage with $\epsilon=0.1$.
We choose top-$100$ block candidates and train their respective networks to verify the best block structure. 
We show top-$2$ block structures in Fig.~\ref{fig:8}(b-c), denoted as \textbf{Block-QNN-A} and \textbf{Block-QNN-B}.
%$\mathcal{B}_a$ and $\mathcal{B}_b$.
% 
% With the block structure, we can construct the network for different datasets and tasks easily.
% 
As shown in Fig.~\ref{fig:8}(a), both top-$2$ blocks are found in the final stage of the Q-learning process, which proves the effectiveness of the proposed method in searching optimal block structures rather than randomly searching a large amount of models.
Furthermore, we observe that the generated blocks share similar properties with those state-of-the-art hand-crafted networks. For example, Block-QNN-A and Block-QNN-B contain short-cut connections and multi-branch structures which have been manually designed in residual-based and inception-based networks. Compared to other auto-generated methods, the networks generated by our approach are more elegant and can automatically and effectively reveal the beneficial properties for optimal network structure.

To squeeze the search space, as stated in Section~\ref{subsec:blocks}, we define a Pre-activation Convolutional Cell (PCC) consists of three components, \ie~ReLU, convolution and batch normalization (BN). We show the superiority of the PCC, searching a combination of three components, in Fig.~\ref{fig:pcc}, compared to the separate search of each component. Searching the three components separately is more likely to generate ``bad'' blocks and also needs more search space and time to pursue ``good'' blocks.

\subsection{Block Connection Analysis}

\begin{figure}[tbp]
	\centering
	\includegraphics[width=\linewidth]{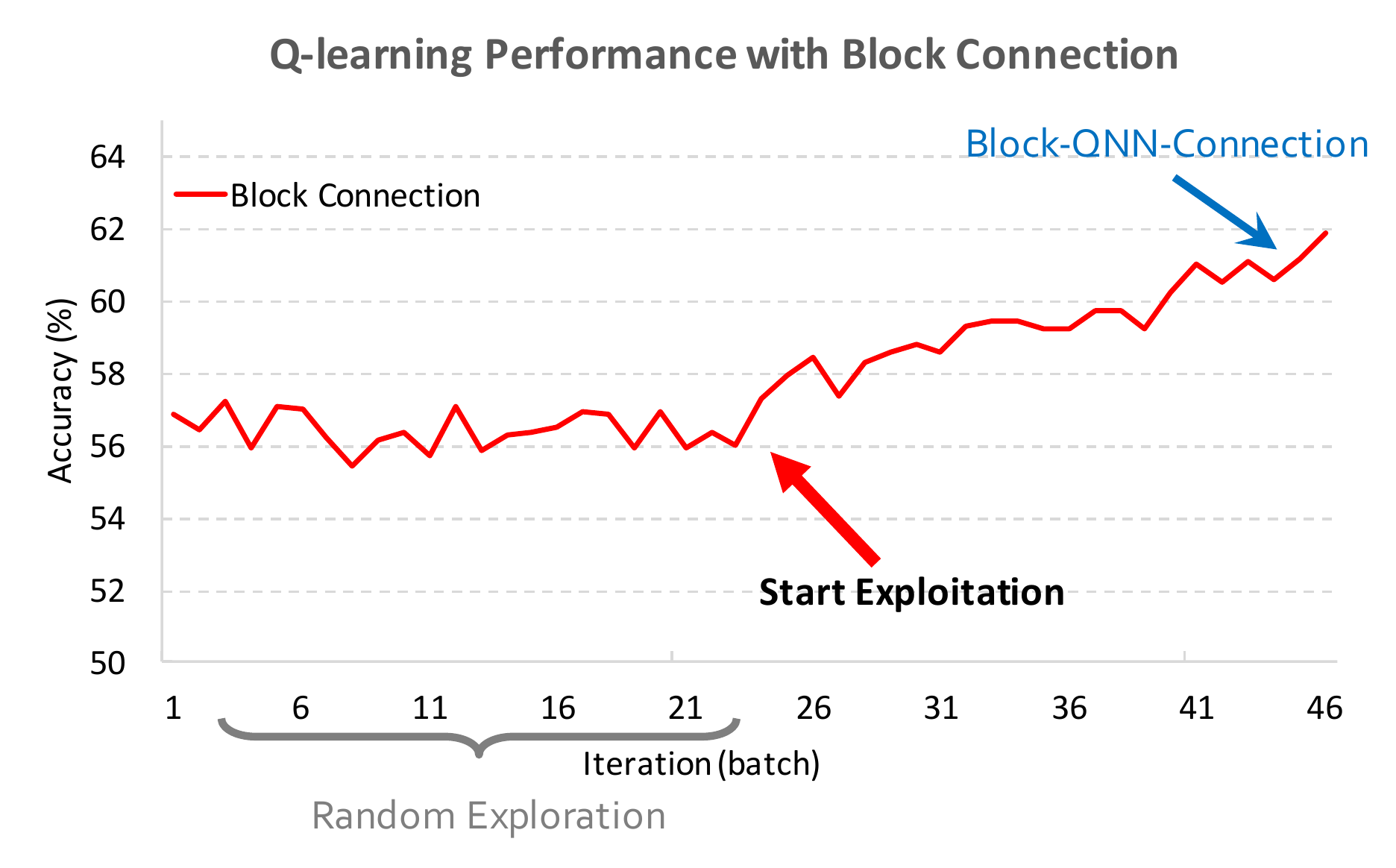}
	\caption{Q-learning performance in block connection generation on CIFAR-$100$. The accuracy goes up with the epsilon decrease and the top model is found in the final stage. The training process is relatively shorter than the standard BlockQNN because the search space of connection is smaller than the block. }\label{fig:con_q}
	%\vspace{-0.3cm}
\end{figure}

\begin{figure*}[tbp]
	\centering
	\includegraphics[width=\linewidth]{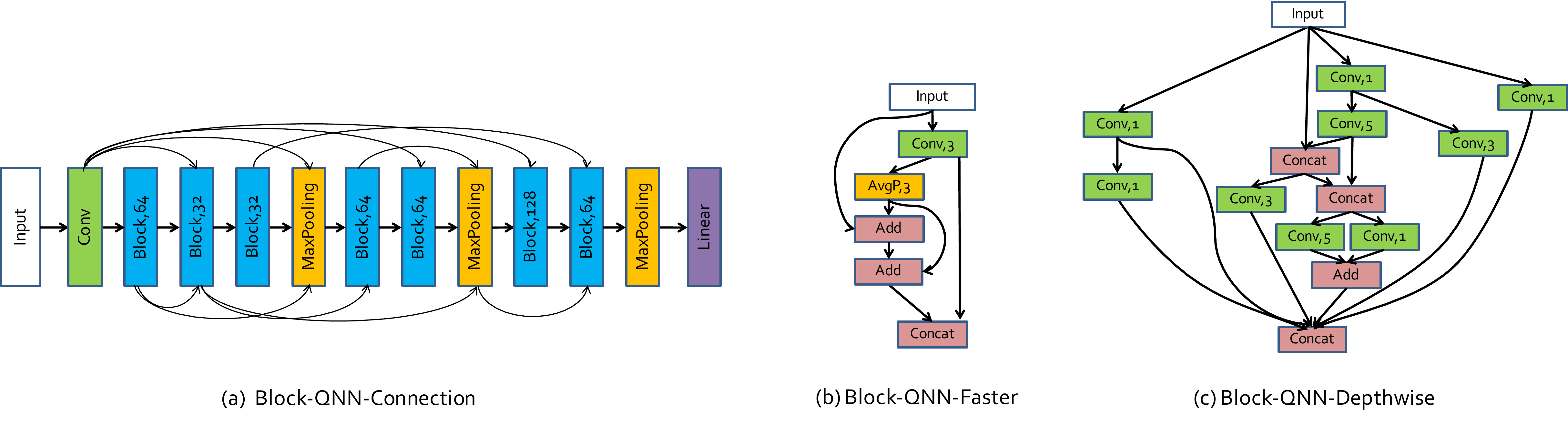}
	%\vspace{0.05cm}
	\caption{(a) Topology of the best connection style between blocks generated by block Connection, named Block-QNN-Connection. Note that the basic block we used here is Block-QNN-S, \ie~Fig.~\ref{fig:8}(d). (b) Topology of the top block structures generated by Faster BlockQNN. We call it Block-QNN-Faster. (c) Topology of the best block structures generated with advanced depthwise convolution operation, named Block-QNN-Depthwise. All convolution operation be replaced by a cell with four components,  \ie~\textit{ReLU}, \textit{Depthwise Convolution}, \textit{Pointwise Convolution} and \textit{Batch Normalization}. 
     }
	\label{fig:block_con_dp_faster}
\end{figure*}

\begin{figure}[tbp]
	\centering
	\includegraphics[width=\linewidth]{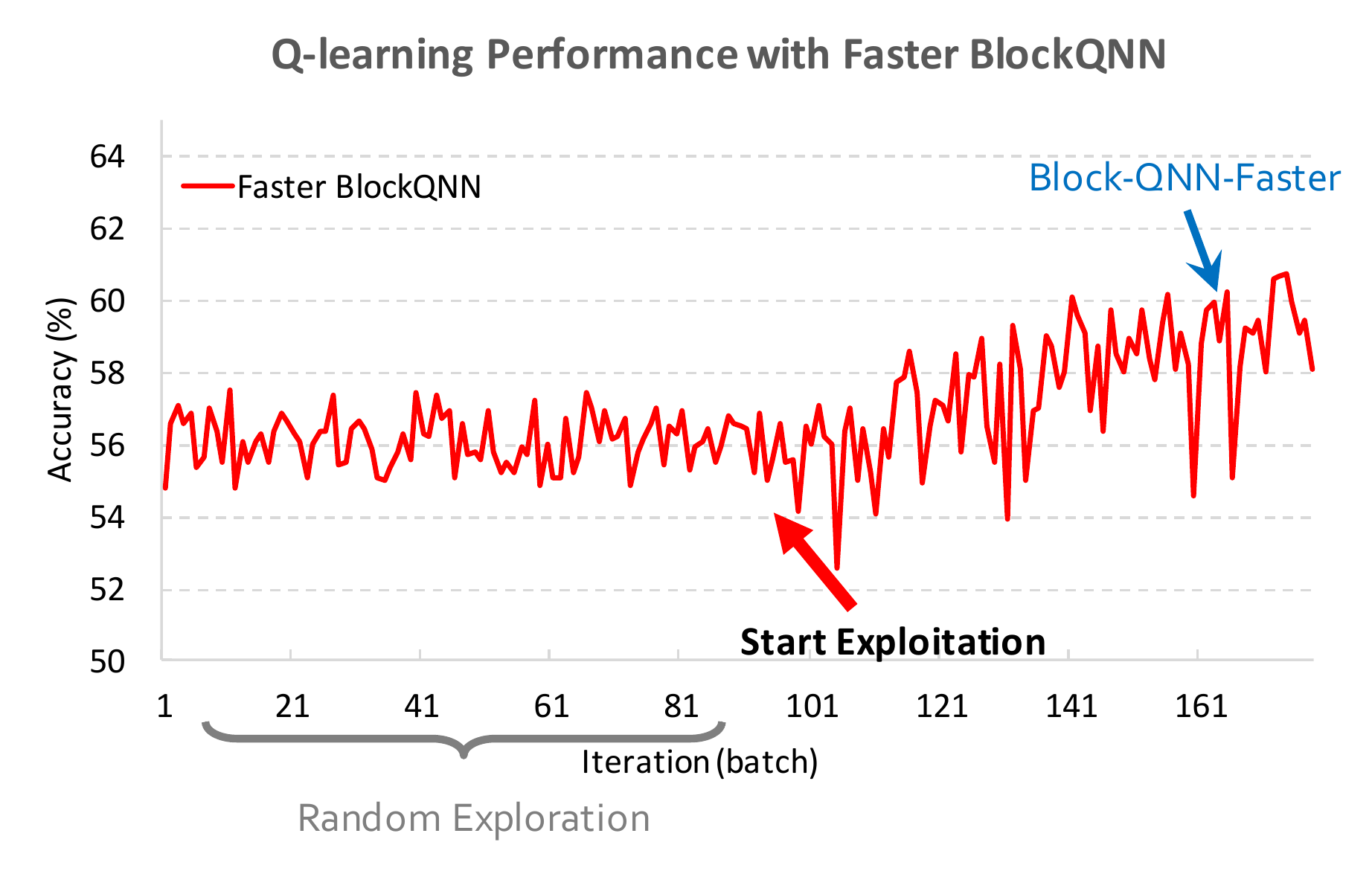}
	\caption{Q-learning performance in faster BlockQNN generation on CIFAR-$100$. The accuracy goes up with the epsilon decrease and the top model is found in the final stage. We can find that the convergent curve of the Q-learning performance grows shakily, it may caused by the errors between Faster BlockQNN predictor and early stop training}\label{fig:faster_q}
	%\vspace{-0.3cm}
\end{figure}

Fig.~\ref{fig:con_q} provides early stop accuracies over $46$ batches on CIFAR-$100$, each of which is averaged over $64$ auto-generated block connection candidates within in each mini-batch. After random exploration, the early stop accuracy grows steadily. The mean accuracy within the period of random exploration is $56\%$ while finally achieves $62\%$ in the last stage with $\epsilon=0.1$. It is shown that the convergent curve of the Q-learning performance saw in the above is similar with Fig.~\ref{fig:8}(a).
The basic block structure we used is Block-QNN-S will be introduced in section~\ref{sec:result_cifar}.
We choose top-$64$ block connection and train their respective networks to verify the best connection style. 
We show the top block connection in Fig.~\ref{fig:block_con_dp_faster}(a), denoted as \textbf{Block-QNN-Connection}.
Moreover, we observe that the generated blocks connection style is totally different from the sequentially connection.
The connection prefer to connect blocks with different resolutions (high resolutions to low resolutions). It may improve the network performance by combining features across multiple resolutions. This conception has been widely used in segmentation area~\cite{long2015fully}, but our method found that it also work well in classification task with appropriate connection style.
The only weakness is that the generated network is dataset-specific that can not transfer to other task with different input data size. Here, the Block-QNN-Connection can only handle image with $32\times32$. The exploration of universal block connection formulation with transferable ability is still a open problem.

\subsection{Faster BlockQNN with 1 GPUs in 20 Hours }

As shown in Fig.~\ref{fig:faster_q}, the mean accuracies over $178$ batches on CIFAR-$100$, each of which is averaged over $64$ auto-generated block-wise network candidates within in each mini-batch. After random exploration, the mean accuracy grows shakily. The mean accuracy within the period of random exploration is $56\%$ while finally achieves $60\%$ in the last stage with $\epsilon=0.1$. We can find that the convergent curve of the Q-learning performance saw in the above is different with Fig.~\ref{fig:8}(a). The mean accuracy is also lower than the standard BlockQNN. It may caused by the errors between Faster BlockQNN predictor and early stop training.
We choose top-$100$ block candidates and train their respective networks to verify the best block structure. As shown in Fig.~\ref{fig:faster_q}, the top block is found in the final stage of the Q-learning process, besides the mean accuracy grows even if the curve is shakily, which proves the effectiveness of the proposed faster method in searching optimal block structures.
We show the top block structures in Fig.~\ref{fig:block_con_dp_faster}(b), denoted as \textbf{Block-QNN-Faster}.
Furthermore, we observe that the top blocks generated by Faster BlockQNN are also different from the top blocks by standard BlockQNN. Block-QNN-Faster doesn't contain multi-branch structure, instead it can be seen as a variant of DenseNet and ResNet. Although the generated block is not complex as the standard BlockQNN blocks, it needs very little computing resources to get comparable performance which is afforded for common deep learning researcher. 
% 

% waiting for depthwise result.....
\subsection{Results on CIFAR}
\label{sec:result_cifar}
Due to the small size of images (\ie~$32\times 32$) in CIFAR, we set block stack number as $N=4$. We compare our generated best architectures with the state-of-the-art hand-crafted networks or auto-generated networks in Table~\ref{table:3}.

\vspace{0.1cm} \noindent \textit{Comparison with hand-crafted networks -}
It shows that our Block-QNN networks (\ie~\textbf{Block-QNN-A} and~\textbf{Block-QNN-B}) outperform most hand-crafted networks. The DenseNet-BC~\cite{huang2016densely} uses additional $1\times 1$ convolutions in each composite function and compressive transition layer to reduce parameters and improve performance, which is not adopted in our design. Our performance can be further improved by using this prior knowledge.

%that we do not introduce extra prior knowledge during training.

\vspace{0.1cm} \noindent \textit{Comparison with auto-generated networks -}
Our approach achieves a significant improvement to the MetaQNN~\cite{baker2016designing}, and even better than NAS's best model (\ie~\textit{NASv3 more filters})~\cite{zoph2016neural} proposed by Google brain which needs an expensive costs on time and GPU resources. As shown in Table~\ref{table:4}, NAS trains the whole system on $\mathbf{800}$ \textbf{GPUs} in $\mathbf{28}$ \textbf{days} while we only need $\mathbf{32}$ \textbf{GPUs} in $\mathbf{3}$ \textbf{days} to get state-of-the-art performance. 
Moreover, the faster version can get a comparable result with only $\mathbf{1}$ \textbf{GPU} in $\mathbf{20}$ \textbf{hours} which is $\mathbf{100}$\textbf{x} less expensive than standard BlockQNN.

\vspace{0.1cm} \noindent \textit{Transfer block from CIFAR-$100$ to CIFAR-$10$ -}
We transfer the top blocks learned from CIFAR-$100$ to CIFAR-$10$ dataset, all experiment settings are the same. As shown in Table~\ref{table:3}, the blocks can also achieve state-of-the-art results on CIFAR-$10$ dataset with $3.60\%$ error rate that proved Block-QNN networks have powerful transferable ability.

\vspace{0.1cm} \noindent \textit{Analysis on network parameters -}
The networks generated by our method might be complex with a large amount of parameters since we do not add any constraints during training. We further conduct an experiment on searching networks with limited parameters and adaptive block numbers. We set the maximal parameter number as $10$M and obtain an optimal block (\ie~\textbf{Block-QNN-S}) which outperforms NASv3 with less parameters, as shown in Fig.~\ref{fig:8}(d). In addition, when involving more filters in each convolutional layer (\eg~from [$32$,$64$,$128$] to [$80$,$160$,$320$]), we can achieve even better result ($2.80\%$).

\vspace{0.1cm} \noindent \textit{Block connection -}
The optimal block connection style we find is much better than the sequentially connection with the basic block structure, \ie~Block-QNN-S, as shown in Table~\ref{table:3} (\ie~from $3.30\%$ to $3.01\%$ with even less parameters). Moreover, when involving more filters in each convolutional layer, we can further improve the performance ($2.35\%$). The empirical evidence indicates that the connection style between blocks can further improve the performance of networks. But as discussed above, the only weakness of block connection is that the generated network is dataset-specific that can not transfer to other task with different input data size.

\vspace{0.1cm} \noindent \textit{Advanced depthwise convolution operation -}
To further reduce the computational complexity of the generated network, we introduce the	advanced depthwise convolution operation. Each convolution operation refers to a cell with four components,  \ie~\textit{ReLU}, \textit{Depthwise Convolution}, \textit{Pointwise Convolution} and \textit{Batch Normalization}. Note that we remove 
Batch Normalization and ReLU between the depthwise and pointwise operations in the cell. After the searching process with depthwise convolution, we obtain an optimal block only with $3.3M$ parameters (\ie~\textbf{Block-QNN-depthwise}) which achieve a result with $2.58\%$ error rate on CIFAR-$10$, as shown in Fig.~\ref{fig:block_con_dp_faster}(c). Moreover, our model outperforms all other networks with similar parameters.

%the model with more filters get a new state of the art on CIFAR with $2.58\%$ error rate.
%outperforms other models

%table 3
\begin{table}[t!]
	%% increase table row spacing, adjust to taste
	\renewcommand{\arraystretch}{1.3}
	%\centering

	\begin{center}
		\caption{Block-QNN's results (error rate) compare with state-of-the-art methods on CIFAR-$10$ (C-$10$) and CIFAR-$100$ (C-$100$) dataset. }\label{table:3}	
		\footnotesize
		\begin{tabular}{c|c|c|c|c}
			\hline
			Method  & Depth &Para &C-10  &C-100 \\
			\hline
			\hline
			%Network in Network~\cite{lin2013network} & - & 8.81 & 35.68 \\
			%\hline
			%Highway Network~\cite{srivastava2015highway} & - & 7.72 & -  \\
			%\hline
			%All-CNN~\cite{springenberg2014striving} & - & 7.25 & 33.71  \\
			%\hline
			VGG~\cite{simonyan2014very} & - & &7.25 & -  \\
			\hline
			\hline
			ResNet~\cite{he2015deep} & 110 &1.7M &6.61 & -  \\
			\hline
			Wide ResNet~\cite{zagoruyko2016wide} & 28 & 36.5M&4.17 & 20.5  \\
			\hline
			ResNet (pre-activation)~\cite{he2016identity} & 1001 &10.2M &4.62 & 22.71 \\
			\hline
			DenseNet (k = 12) ~\cite{huang2016densely} & 40 & 1.0M&5.24 & 24.42 \\
			DenseNet (k = 12)~\cite{huang2016densely} & 100 &7.0M &4.10 & 20.20 \\
			DenseNet (k = 24)~\cite{huang2016densely} & 100 &27.2M &3.74 & 19.25 \\
			DenseNet-BC (k = 40)~\cite{huang2016densely} & 190 &25.6M &3.46 & 17.18  \\
			\hline
			\hline
			MetaQNN (ensemble)~\cite{baker2016designing} & - & - &7.32 & - \\
			MetaQNN (top model)~\cite{baker2016designing} & - & 11.2M &6.92 & 27.14 \\
			\hline
			NAS v1~\cite{zoph2016neural} & 15 & 4.2M&5.50 & - \\
			NAS v2~\cite{zoph2016neural} & 20 &2.5M &6.01 & - \\
			NAS v3~\cite{zoph2016neural} & 39 &7.1M &4.47 & - \\
			NAS v3 more filters~\cite{zoph2016neural} & 39 &37.4M &3.65 & - \\
			\hline
			NASNet-A (6 @ 768)~\cite{zoph2018learning}&-&3.3M&2.65&-\\
			\hline
			Block-QNN-A, N=4~\cite{zhong2018blockqnn} & 25 &-&3.60 &18.64  \\
			Block-QNN-B, N=4~\cite{zhong2018blockqnn} & 37 &-&3.80 & 18.72\\
			Block-QNN-S, N=2 & 19 &6.1M & 3.30 & 17.05 \\
			Block-QNN-S more filters &22&39.8M&2.80&15.56\\
			Block-QNN-Faster  &25&3.9M&3.57&18.21  \\
			Block-QNN-Faster more filters &25&34.4M&3.15&16.74  \\
			Block-QNN-Connection &22&3.9M&3.01&16.07\\
			Block-QNN-Connection more filters &22&33.3M&2.35&14.83\\
			Block-QNN-Depthwise, N=3 &19&3.3M&2.58&15.28  \\
			%Block-QNN-Depthwise more filters &22&39.8M&None&18.06  \\
			\hline
		\end{tabular}
	\end{center}

	%\vspace{-0.3cm}
\end{table}

\subsection{Transfer to ImageNet}
To demonstrate the generalizability of our approach, we transfer the block structure learned from CIFAR to ImageNet dataset. 

For the ImageNet task, we set block repeat number $N=3$ and add more down sampling operation before blocks, the filters for convolution layers in different level blocks are [$64$,$128$,$256$,$512$]. We use the best blocks structure learned from CIFAR-$100$ directly without any fine-tuning, and the generated network initialized with MSRA initialization as same as above. The experimental results are shown in Table~\ref{table:5}. The network generated by our framework can get competitive result compared with other human designed models. The recently proposed methods such as Xception~\cite{chollet2016xception} and ResNext~\cite{xie2016aggregated} use special depthwise convolution operation to reduce their total number of parameters and to improve performance. In our work, the block structures with depthwise convolution operation, \ie~\textbf{Block-QNN-Depthwise}, can outperform all these hand-crafted networks. 
SENet~\cite{hu2018senet} use fully connection layer to recalibrate channel-wise feature responses which we do not adopt this operation in our search space. Besides, SENet can be seen as a Plug-in block to any  backbone network, we will consider this in our future work to further improve the performance.

%table 4
\begin{table}[t!]
	%% increase table row spacing, adjust to taste
	\renewcommand{\arraystretch}{1.3}
	%\centering
	\footnotesize
	\begin{center}
		\caption{The required computing resource and time of our approach compare with other automatic designing network methods.}\label{table:4}
		\begin{tabular}{c|c|c|c}
			\hline
			Method  & Best Model on CIFAR10  &GPUs & Time(days)\\
			\hline
			
			\hline
			MetaQNN~\cite{baker2016designing} & 6.92 & 10 & 10 \\
			\hline
			NAS~\cite{zoph2016neural} & 3.65 & 800 & 28  \\
			\hline
			NASNet-A~\cite{zoph2018learning} & 2.65 & 450 & 3-4  \\
			\hline
			
			\hline
			BlockQNN & 2.80 & 32 & 3  \\
			Faster BlockQNN & 3.15 & 1 & 0.8  \\
		    Depthwise BlockQNN & 2.58 & 32 & 3  \\
		    Connection BlockQNN & 2.35 & 32 & 1  \\
			\hline
		\end{tabular}
	\end{center}

	%\vspace{-0.1cm}
\end{table}

%table 5
\begin{table}[t!]
	%% increase table row spacing, adjust to taste
	\renewcommand{\arraystretch}{1.3}
	%\centering
	\footnotesize
	\begin{center}
		\caption{Block-QNN's results (single-crop error rate) compare with modern methods on ImageNet-$1$K Dataset.}\label{table:5}
		\begin{tabular}{c|c|c|c|c}
			\hline
			Method  & Input Size &Para & Top-1& Top-5\\
			\hline
			
			\hline
			VGG~\cite{simonyan2014very} & x224 & 138M & 28.5 & 9.90  \\
			\hline
			Inception V1~\cite{szegedy2015going} & x224 & 7M& 30.2 & 10.10  \\
			%\hline
			Inception V2~\cite{ioffe2015batch} & x224 & 11M& 25.2 & 7.80  \\
			
			ResNet-50~\cite{he2016identity}  & x224& 26M &  24.7 & 7.80 \\
			%\hline
			ResNet-152~\cite{he2016identity} & x224&  60M &23.0 & 6.70 \\
			
			Xception(our test) & x224& 23M &  23.6 & 7.10 \\
			%\hline
			ResNext-101(64x4d)~\cite{xie2016aggregated} & x224&  84M & 20.4 & 5.30 \\
			DPN-131~\cite{chen2017dual}& x224&  80M & 19.93 & 5.12 \\
			\hline
			Xception~\cite{chollet2016xception}  & x299& 23M &  21.00 & 5.50 \\
			Inception-resnet-v2~\cite{szegedy2017inception} & x299& 56M &  19.90 & 4.90 \\
			Very Deep Inception-ResNet~\cite{zhang2017polynet}& x299&  130M & 19.10 & 4.48 \\
			PolyNet~\cite{zhang2017polynet}& x331&  92M & 18.71 & 4.25 \\
			DPN-131~\cite{chen2017dual}& x320&  80M & 18.55 & 4.16 \\			
			\hline
			NASNet-A(6 @ 4032)(our test) & x224 & 89M  &19.90  & 5.27  \\
			NASNet-A(6 @ 4032)~\cite{zoph2018learning}& x331 & 89M  &17.30  & 3.80  \\
			\hline
			
			\hline
			%Block-QNN-A  & 224x224 &  &  \\
			Block-QNN-B, N=3~\cite{zhong2018blockqnn} & x224 & -  &24.3  & 7.40  \\
			%Block-QNN-S, N=3 & 224x224 & 38  &22.9  & 6.56  \\
			Block-QNN-S, N=3 & x224 & 95M  &21.9  & 6.16  \\
			Block-QNN-Depthwise & x224 & 91M  &19.00  & 4.58  \\
			\hline
			Block-QNN-Depthwise & \tabincell{l}{train x224 \\ test\ \  x320}    & 91M  &18.00  & 4.00  \\
			\hline
			%Block-QNN-Depthwise, N=3 & x331   & 38  &None  & None  \\
			%\hline			
		\end{tabular}
	\end{center}
	%\vspace{0.2cm}

	%\vspace{-0.3cm}
\end{table}

%%table 6
%\begin{table}[t!]
%	%% increase table row spacing, adjust to taste
%	\renewcommand{\arraystretch}{1.3}
%	%\centering
%	\footnotesize
%	\begin{center}
%		\caption{Block-QNN's results (single-crop error rate) compare with mobile model on ImageNet-$1$K Dataset. All models use 224x224 images.}\label{table:6}
%		\begin{tabular}{c|c|c|c}
%			\hline
%			Method  & Paramters  & Top-1& Top-5\\
%			\hline
%			
%			MobileNet & 4.2M& 29.4 & 10.5 \\
%			ShuffleNet (2x) & 5M&  29.1 & 10.2 \\			
%			\hline
%			%NASNet-A(4 @ 1056) & 5.3M & 564M  &26.0  & 8.4  \\
%			%\hline
%			
%			\hline
%			Block-QNN-Depthwise & 5.3M &26.8  & 9.9 \\
%			\hline			
%		\end{tabular}
%	\end{center}
%	%\vspace{0.2cm}
%	
%	%\vspace{-0.3cm}
%\end{table}

\begin{figure*}[tbp]
	\centering
	\includegraphics[width=\linewidth]{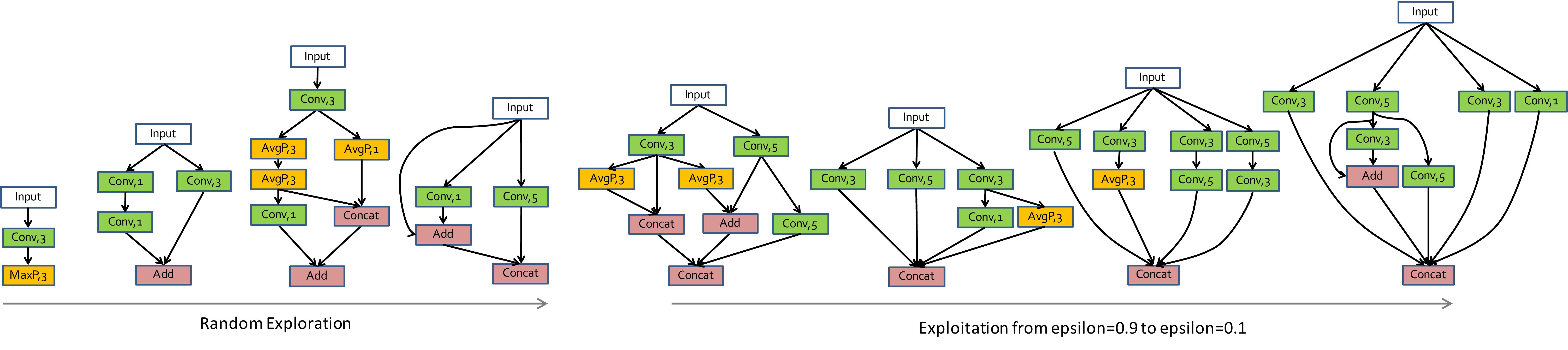}
	\caption{Evolutionary process of blocks generated by BlockQNN. We sample the block structures with median performance at iteration $[1, 30, 60, 90, 110, 130, 150, 170]$ to compare the difference between the blocks in the random exploration stage and the blocks in the exploitation stage.}
	\label{fig:3}
\end{figure*}

\begin{figure*}[tbp]
	\centering
	\includegraphics[width=\linewidth]{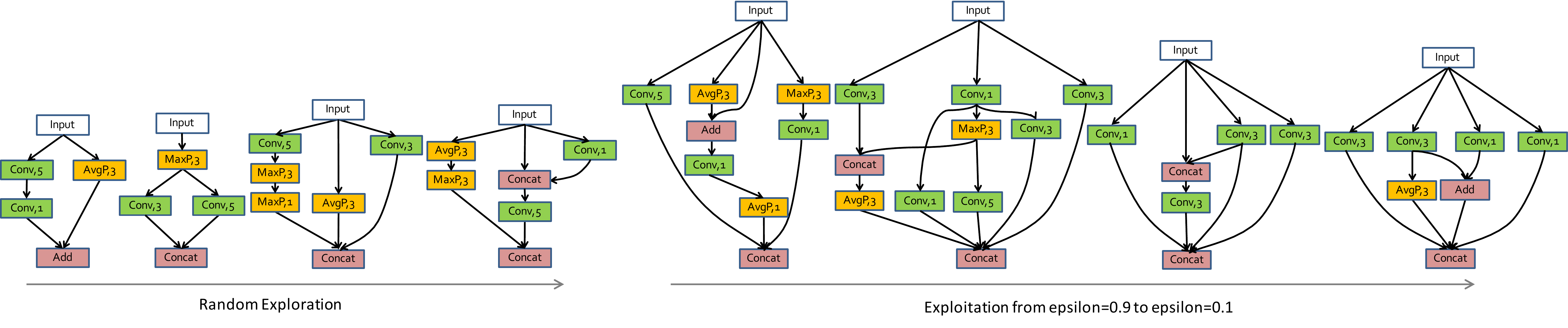}
	\caption{Evolutionary process of blocks generated by BlockQNN with limited parameters and adaptive block numbers (BlockQNN-L). We sample the block structures with median performance at iteration $[1, 30, 60, 90, 110, 130, 150, 170]$ to compare the difference between the blocks in the random exploration stage and the blocks in the exploitation stage.}
	\label{fig:4}
\end{figure*}

As far as we known, most previous works of automatic network generation did not report competitive result on large scale image classification datasets. With the conception of block learning, we can transfer our architecture learned in small datasets to big dataset like ImageNet task easily.
Furthermore, the model Block-QNN-Depthwise achieves a surprising performance on ImageNet ($82.0\%$ Top1 $96.0\%$ Top5) when trained on 224x224 and tested on 320x320 images based on the single crop and single model condition. With limited compute resource, we can not train our model on high resolution images (~\ie 331x331) directly like NASNet~\cite{zoph2018learning}. For a fair comparison, we train Block-QNN-Depthwise and NASNet-A(6 @ 4032) on 224x224 images with same parameter setting. As can be seen from Table~\ref{table:5}, our Block-QNN-Depthwise reduces the top-1 error rate by an absolute value of 0.9\% compared with the NASNet-A(6 @ 4032). Moreover, the inference speed of Block-QNN-Depthwise (281 Image/Second) is around 30\% faster than NASNet-A (222 Image/Second) tested on same computing platform and hardware environment (PyTorch and 8x NVIDIA 1080Ti). The experimental result on ImageNet shows that our auto-generated model can achieve very competitive performances compare with state-of-the-art models.

%At last, we show how well the best block may perform in a resource-constrained setting for mobile devices in Table~\ref{table:6}. MobileNet~\cite{howard2017mobilenets} and ShuffleNet~\cite{xiangyu2017shufflenet} get state-of-the-art results on 224x224 images with nearly 5M parameters. Our Block-QNN-Depthwise achieve a leading performance,~\ie 26.8\%, outperform previous models but with comparable model size. The experimental result on ImageNet shows that our auto-generated model can achieve very competitive performances on both large and small models.

%\vspace{-0.3cm}

\subsection{Evolutionary Process of Auto-Generated Blocks}

We sample the block structures with median performance generated by our approach in different stage, \ie~at iteration $[1, 30, 60, 90, 110, 130, 150, 170]$, to show the evolutionary process. As illustrated in Figure~\ref{fig:3} and Figure~\ref{fig:4}, \ie~BlockQNN and BlockQNN-L respectively, the block structures generated in the random exploration stage is much simpler than the structures generated in the exploitation stage.

In the exploitation stage, the multi-branch structures appear frequently. Note that the connection numbers is gradually increase and the block tend choose "Concat" as the last layer. And we can find that the short-cut connections and elemental add layers are common in the exploitation stage. Additionally, blocks generated by BlockQNN-L have less "Conv,$5$" layers, \ie~convolution layer with kernel size of $5$, since the limitation of the parameters.

These prove that our approach can learn the universal design concepts for good network blocks. Compare to other automatic network architecture design methods, our generated networks are more elegant and model explicable.

\subsection{Efficiency of BlockQNN}

We demonstrate the effectiveness of our proposed BlockQNN on network architecture generation on the CIFAR-$100$ dataset as compared to random search given an equivalent amount of training iterations, \ie~number of sampled networks. We define the effectiveness of a network architecture auto-generation algorithm as the increase in top auto-generated network performance from the initial random exploration to exploitation, since we aim to getting optimal auto-generated network instead of promoting the average performance.

\begin{figure}[t!]
	\centering
	\includegraphics[width=\linewidth]{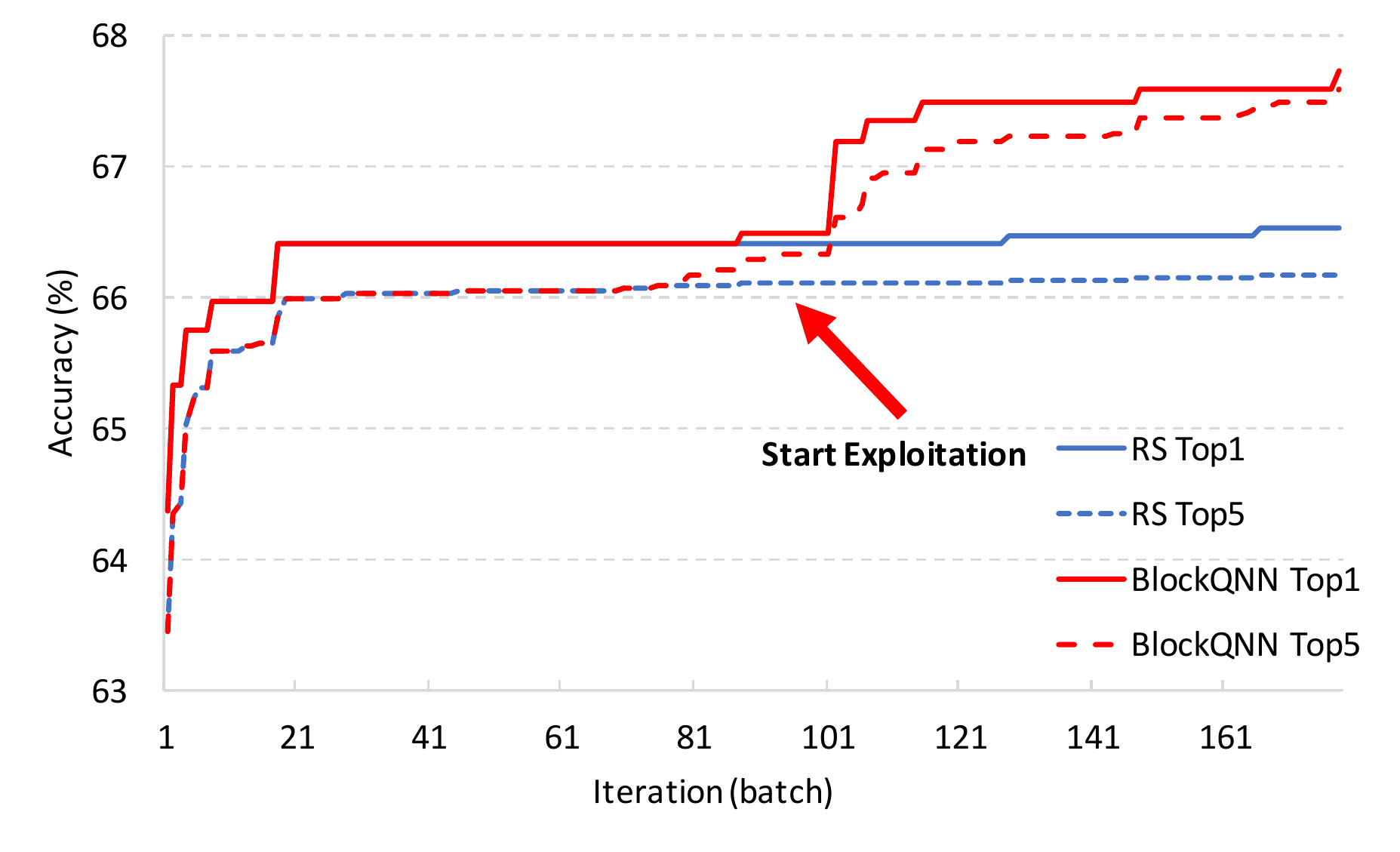}
	\caption{Measuring the efficiency of BlockQNN to random search (RS) for learning neural architectures. The x-axis measures the training iterations (batch size is $64$), \ie~total number of architectures sampled, and the y-axis is the early stop performance after $12$ epochs on CIFAR-$100$ training. Each pair of curves measures the mean accuracy across top ranking models generated by each algorithm. Best viewed in color.}
	\label{fig:1}
	\vspace{-0.1cm}
\end{figure}

Figure~\ref{fig:1} shows the performance of BlockQNN and random search (RS) for a complete training process, \ie~sampling $11, 392$ blocks in total. We can find that the best model generated by BlockQNN is markedly better than the best model found by RS by over $1\%$ in the exploitation phase on CIFAR-$100$ dataset. We observe this in the mean performance of the top-$5$ models generated by BlockQNN compares to RS. Note that the compared random search method start from the same exploration phase as BlockQNN for fairness.

\begin{figure}[t!]
	\centering
	\includegraphics[width=\linewidth]{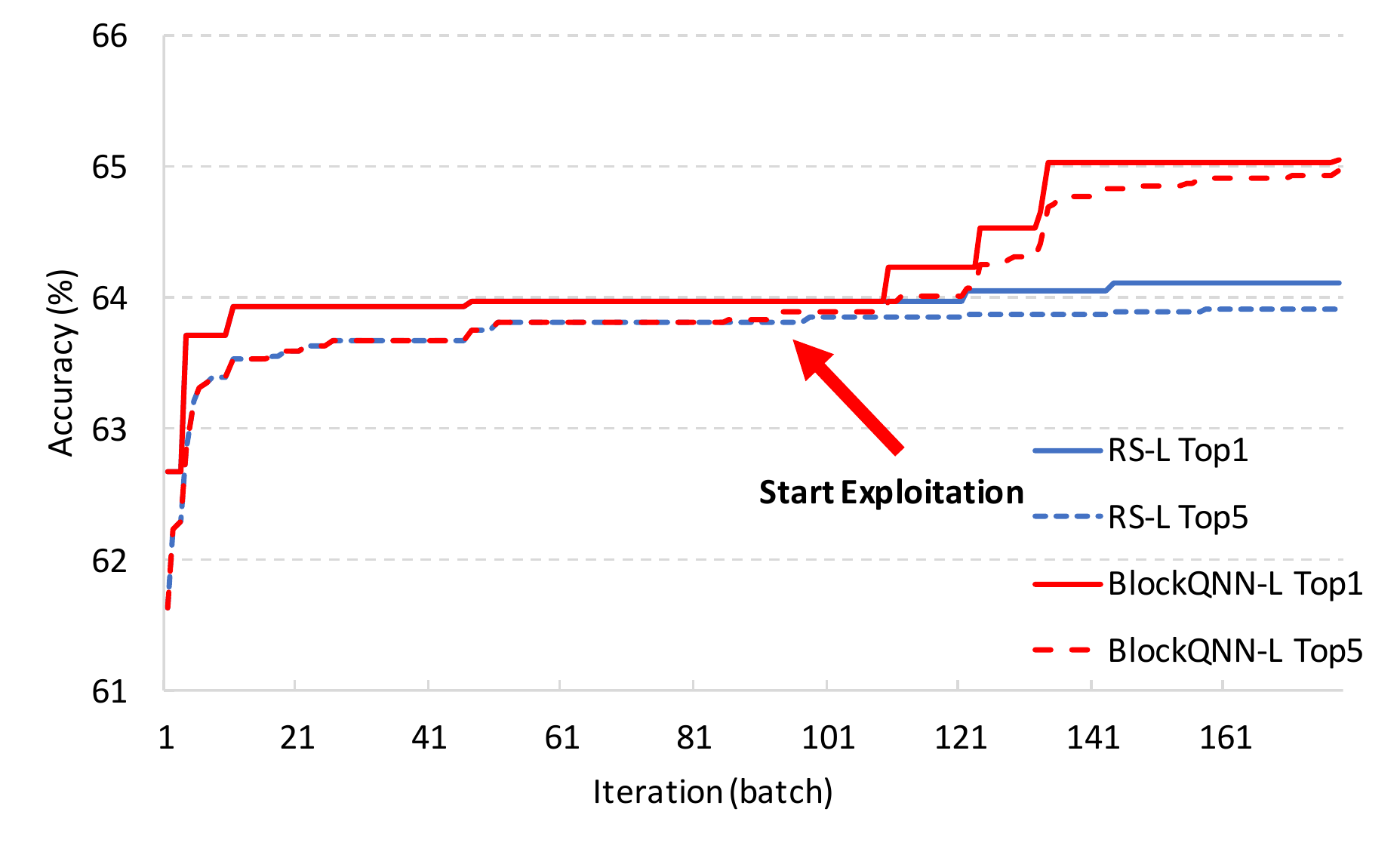}
	\caption{Measuring the efficiency of BlockQNN with limited parameters and adaptive block numbers (BlockQNN-L) to random search with limited parameters and adaptive block numbers (RS-L) for learning neural architectures. The x-axis measures the training iterations (batch size is $64$), \ie~total number of architectures sampled, and the y-axis is the early stop performance after $12$ epochs on CIFAR-$100$ training. Each pair of curves measures the mean accuracy across top ranking models generated by each algorithm. Best viewed in color.}
	\label{fig:2}
	\vspace{-0.1cm}
\end{figure}

Figure~\ref{fig:2} shows the performance of BlockQNN with limited parameters and adaptive block numbers (BlockQNN-L) and random search with limited parameters and adaptive block numbers (RS-L) for a complete training process. We can see the same phenomenon, BlockQNN-L outperform RS-L by over $1\%$ in the exploitation phase. These results prove that our BlockQNN can learn to generate better network architectures rather than random search.

%------------------------------------------------------------------------

\subsection{Additional Experiment}
We also use BlockQNN to generate optimal model on person key-points task. The training process is conducted on MPII dataset, and then, we transfer the best model found in MPII to COCO challenge. It costs $5$ days to complete the searching process. The auto-generated network for key-points task outperform the state-of-the-art hourglass $2$ stacks network, \ie~$70.5$ AP compares to $70.1$ AP on COCO validation dataset.

\section{Conclusion and future work}
In this paper, we showed how to efficiently design high performance network blocks with Q-learning, with a distributed asynchronous Q-learning framework and an early stop strategy for fast block structures search. We applied the framework to automatic block generation for constructing good convolutional network. Our Block-QNN networks outperform modern hand-crafted networks and other auto-generated networks in image classification tasks. The best block structure, which achieves a state-of-the-art performance on CIFAR, can be transferred to the large-scale dataset ImageNet easily, and also yield a very competitive performance compared with best hand-crafted networks. We showed that searching with the block design strategy can get more elegant and model explicable network architectures. 
Furthermore, we propose a faster version which is $100$x less expensive than standard BlockQNN and still give a comparable result. The proposed framework makes it possible for common deep learning researchers to join automated neural network design on limited computing resource.
We also discussed the different connection style between blocks, and the empirical results showed that sequentially connection is not the optimal style.

This work highlights the general trend from hand-crafted networks to auto-generated networks in deep learning community. In the future, we will try to automatic generate backbone network for other task directly such as detection, segmentation and tracking. For these task, they often use classification network as an backbone, which may be not the optimal choice. Another important future research topic is the automatically preprocessing the input data, designing the network and setting the training hyperparameters in an unified framework.
%We believe our work can free the human from complex network architecture design and tuning work in the future.

% or


%\appendices

%\section{no appendix here for now}

\ifCLASSOPTIONcompsoc
  \section*{Acknowledgments}
  This work has been supported by the National Natural Science Foundation of China (NSFC) Grants 61721004 and 61633021. 
\else
  \section*{Acknowledgment}
\fi


\ifCLASSOPTIONcaptionsoff
  \newpage
\fi



\bibliographystyle{IEEEtran}
\bibliography{egbib.bib}

\end{document}